\documentclass[fleqn,11pt]{article}
\usepackage[a4paper, margin=3cm]{geometry}
\usepackage[utf8]{inputenc}
\usepackage[T1]{fontenc}
\usepackage{subcaption}
\usepackage{xcolor} 
\usepackage{soul}
\usepackage{graphicx}
\usepackage{booktabs}
\definecolor{color11}{HTML}{4fb688}
\definecolor{color21}{HTML}{005622}
\definecolor{color31}{HTML}{f4fbfc}
\definecolor{color41}{HTML}{70c6ac}
\definecolor{color51}{HTML}{f7fcfd}
\definecolor{color61}{HTML}{2b9452}
\definecolor{color71}{HTML}{61bf9e}
\definecolor{color81}{HTML}{e3f4f7}
\definecolor{color91}{HTML}{005020}
\definecolor{color101}{HTML}{00441b}
\usepackage{multicol}
\usepackage{multirow}
\usepackage{hyperref}
\usepackage{amsmath}
\usepackage{fancyhdr}
\usepackage[shortlabels]{enumitem}

\definecolor{first_color}{HTML}{69b469}
\definecolor{second_color}{HTML}{2d962d}
\definecolor{third_color}{HTML}{f3f9f3}
\definecolor{fourth_color}{HTML}{b4d9b4}
\definecolor{fifth_color}{HTML}{fbfdfb}
\definecolor{sixth_color}{HTML}{118811}
\definecolor{seventh_color}{HTML}{4aa44a}
\definecolor{eighth_color}{HTML}{cde6cd}
\definecolor{ninth_color}{HTML}{2c952c}
\definecolor{tenth_color}{HTML}{a6d2a6}
\definecolor{eleventh_color}{HTML}{e3f1e3}

\usepackage{titlesec}
\titleformat{\subsubsection}
  {\normalfont\itshape}{\thesubsubsection}{1em}{}

\title{\textbf{Fully automatic extraction of morphological traits from the Web: Utopia or reality?}}

\begin{document}

\pagestyle{fancy}
\fancyhead{} 
\fancyhead[RH]{Marcos \emph{et al.} - Fully automatic trait extraction, \thepage}

{
\centering\Large
\textbf{Fully automatic extraction of morphological traits from the Web: utopia or reality?}
}\\
Diego Marcos$^{1,2}$, 
Robert van de Vlasakker$^{5}$,
Ioannis N. Athanasiadis$^{5}$,
Pierre Bonnet$^{4}$,
Hervé Goeau$^{4}$,
Alexis Joly$^{1,2}$,
W. Daniel Kissling$^{6}$,
César Leblanc$^{1,3}$, 
André S.J. van Proosdij$^{5}$,
Konstantinos P. Panousis$^{1,2,7}$\\
$^1$ Inria, Montpellier, France\\
$^2$ University of Montpellier, France\\
$^3$ LIRMM, University of Montpellier, France\\
$^4$ AMAP, Univ. Montpellier, Cirad, Cnrs, Inrae, Ird, Montpellier, France\\
$^5$ Wageningen University \& Research, Wageningen, The Netherlands\\
$^6$ University of Amsterdam, The Netherlands\\
$^7$ Department of Statistics, Athens University of Economics and Business, Greece\\

\begin{abstract}
\noindent \textbf{Premise:} Plant morphological traits, their observable characteristics, are fundamental to understand the role played by each species within their ecosystem.
However, compiling trait information for even a moderate number of species is a demanding task that may take experts years to accomplish.
At the same time, species descriptions on the Web contain massive amounts of information about morphological traits, although the lack of structure makes this source of data impossible to use at scale.\\
\textbf{Method:} To overcome this, we propose to  leverage recent advances in large language models (LLMs) and devise a mechanism for gathering and processing information on plant traits in the form of unstructured textual descriptions, without manual curation.\\
\textbf{Results:} We evaluate our approach by automatically replicating three manually created species-trait matrices. Our method managed to find values for over half of all species-trait pairs, with an F$_1$-score of over 75\%.\\
\textbf{Discussion:}
Our results suggest that large-scale creation of structured trait databases from unstructured online text is now feasible, thanks to the information extraction capabilities of large language models (LLMs). However, the process is currently limited by the availability of textual descriptions that cover all traits of interest.
\end{abstract}

\textbf{Keywords:} Automatic trait extraction; Large language models; Morphological trait matrices; Natural language processing.

\section*{INTRODUCTION}

Traits are observable characteristics of organisms that can be used to answer a variety of questions about their ecology, evolution, and even usefulness to humans.
Morphological traits in particular, \emph{i.e.}, those that correspond to the physical appearance of the organisms, such as the number and color of flower petals, the size and shape of the fruits or the leaf arrangement, are the main cues that humans have been using to identify species for centuries.
However, the sheer number of known species, the variety of morphological traits and the complexity of trait-based descriptions make it extremely challenging to design a comprehensive framework for trait-based descriptions that would be suitable across taxonomic groups. Within this context, recent efforts advocate for  a standard vocabulary to make trait databases cross-compatible~\cite{schneider2019towards} and an open science initiative to leverage the collective effort of the community~\cite{gallagher2020open}.
Nonetheless, this complexity has resulted in most existing databases of traits being limited either in terms of geographic~\cite{falster2021austraits} or taxonomic scope~\cite{kissling2019palmtraits}. Moreover, large community efforts such as TRY~\cite{kattge2011try}, BIEN~\cite{maitner2018bien} or TraitBank~\cite{caldwell2014using}, which aim at covering all plant species, are far from being comprehensive or representative~\cite{kattge2020try}, even if they have amassed millions of contributed trait measurements.
For instance, in TRY version 6, 27 of the 30 species with the highest number of traits are from Western Europe, and three from North America, showcasing a common imbalance in which relatively less data is available for species from biodiverse tropical regions. On the other hand, over 80\% of plant species in TRY have 10 traits or fewer~\cite{kattge2020try}.

At the same time, taxonomists have been carefully categorizing and describing traits for the purpose of species identification since the dawn of taxonomy and, more recently, using them for this task with modern machine learning approaches~\cite{almeida2020not}.
Many of these trait-based descriptions, capturing a vast expertise in different languages and with varying vocabularies, along with large amounts of trait data, can now be found online in the form of textual descriptions.
However, these data do not come in a structured, ready-to-process format, requiring a thorough and laborious curation process in order to render it usable~\cite{endara2018extraction,folk2023floratraiter}. For instance, \cite{coleman2023workflow} estimated between 8 and 23 hours of manual work required \emph{per trait}, using a partially automated workflow on up to 25k Australian taxa, with most of the time being required for manual verification.  \cite{domazetoski2023using} aimed at reducing the reliance on manual labor by training a natural language processing (NLP) model to output the trait values for a limited number of traits, those the model has been trained on, when provided with a textual description. This process supersedes the need for manual work with the need for structured trait information for training, which explains why the authors limited the approach to eight traits.

In this work, we explore the potential of leveraging these morphological descriptions, in the form of text, for filling in gaps in structured trait databases.
We posit that recent advances in NLP models, and particularly large language models (LLMs), bring us closer to exploiting this knowledge in an automated manner.
LLMs have been shown to behave as remarkable zero-shot learners~\cite{kojima2022large}; this  means that they can be leveraged to solve tasks without a single training example via the use of textual instructions in natural language. Among these tasks, LLMs have been shown to excel at the extraction of structured information from text~\cite{wei2023zero}. To this end, we investigate the feasibility of a workflow that, given the names of the species of interest, along with the traits and possible trait values we are considering, fills in a species-trait matrix using web crawling and LLMs. 
This is in contrast to other related approaches for plant trait extraction, which require manual input for either post-processing~\cite{endara2018extraction,coleman2023workflow,folk2023floratraiter}, or preparing a training set~\cite{domazetoski2023using}.

\section*{METHODS}

\begin{figure}[h]
    \centering
    \includegraphics[width =\textwidth]{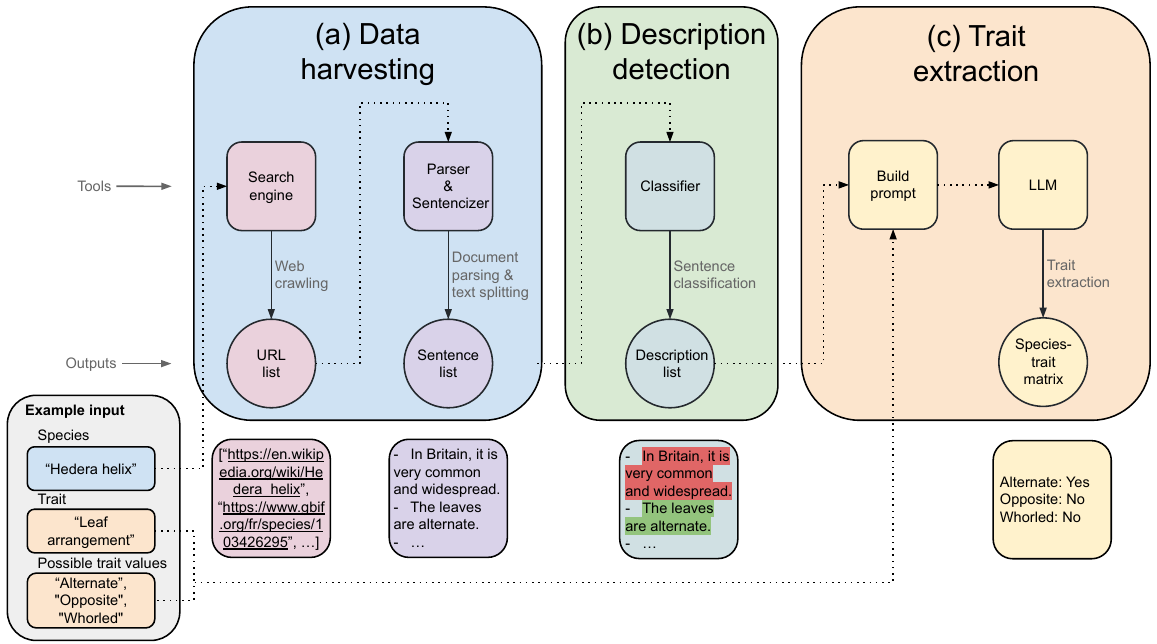}
    \caption[Overview of the methodology]{Overview of the methodology. The panels display the sequence of tasks performed during each of the three main stages: (a) data harvesting, (b) description detection, and (c) trait extraction. Below each task is an example of what its output may look like.}
    \label{fig:workflow}
\end{figure}

We propose a novel framework that only requires three inputs: (i) a list of species of interest, (ii) a list of categorical traits of interest and, for each trait, (iii) a list with all the possible values each trait is allowed to take.
The selected traits should correspond to those typically used in plant species descriptions.
The output is a species-trait table that indicates, for each species, which trait values pertain to it.
Specifically, the workflow (see Figure~\ref{fig:workflow} can be divided in the following steps:
\begin{enumerate}[a)]
\setlength\itemsep{0.1em}
\item \textit{Textual data harvesting}: A search engine API is used to retrieve URLs that are relevant to the species name and downloads the text content therein.
\item \textit{Description detection}: In order to filter out irrelevant text, a binary classification NLP model is used to detect description sentences within the retrieved text.
\item \textit{Trait information extraction}: An LLM is then used to detect all possible categorical trait values within the descriptive text.
\end{enumerate}
%

\subsection*{Species-trait datasets for evaluation}

In order to be able to evaluate the automatic trait extraction workflow, we fix the species, traits, and trait values to those found in three manually created species-trait matrices.
Specifically, we used the following databases:
\begin{itemize}
\setlength\itemsep{0.1em}
\item \textit{Caribbean}: 42 woody species in the Dutch Caribbean, created in the context of this work. It contains 24 traits, with an average of 8.5 possible values per trait (minimum of 2 and a maximum of 22).
\item \textit{West Africa}~\cite{bonnet2005graphic}: 361 species of trees in the West African savanna. We consider all 23 traits, averaging 5.8 possible values per trait (minimum of 2 and a maximum of 10).
\item \textit{Palms}~\cite{kissling2019palmtraits}: We use the 333 species that have a complete trait description. We consider the six categorical traits in the dataset, with an average of 9.5 possible values per trait (minimum of 2 and a maximum of 31).
\end{itemize}

\subsection*{Textual data harvesting and description detection}
\label{sec:harvest}

\subsubsection*{Textual data harvesting}

Given each species of interest, we used the Google Search API and submitted a query with the binary scientific name of the species, in quotes, to make sure that the search engine only returns websites containing the exact species name.
The first 20 returned URLs are then visited and the text scrapped; only HTML sites are considered in this work. We double-check that the species name is present in the HTML header, in order to filter out web pages that are not specifically dedicated to it.
Since the text obtained in this way is unstructured and a large part of it does not correspond to morphological descriptions, we select the sentences most likely to be part of a description by using a custom text classifier, described in the following section.
The full list of Internet domains contributing to the harvested text can be found in the GitHub repository.

\subsubsection*{Description detection}
\label{sec:descr}

We start by formulating an approach for distinguishing between descriptive and non-descriptive sentences in the form of an NLP binary classification task, aimed at filtering out all the text from the retrieved websites that does not describe the morphology of the species.
For instance, in the English Wikipedia page referring to \textit{Hedera helix}, the sentence ``The fruit are purple-black to orange-yellow berries'', would be considered descriptive because it explicitly describes morphological traits, and indeed stems from the ``Description'' section. We are interested in such sentences from which trait values can potentially be extracted.
On the other hand, the sentence ``Once ivy is established it is very difficult to control or eradicate'', from the ``Control and eradication'' section, does not explicitly describe any morphological traits, and is thus considered non-descriptive.
Given a sentence, we need an automated approach to determine if said sentence is descriptive or not.
Such a model can be trained without the need for manual annotations by leveraging structured online sources, such as Wikipedia, in which a ``Description'' section is often present and can be used to obtain descriptive training samples, while text from the other sections can be used as non-descriptive samples.
This model can in turn be used to collect descriptive sentences from other, less structured but relevant, websites for further processing.

\paragraph{\underline{\normalfont{Creation of the training dataset.}}}
We first need to create a large dataset of descriptive and non-descriptive text via parsing structured websites. To this end, we select four different web sources that: (i) comprise large databases, and (ii) have rich scholarly content about species descriptions, namely:
\begin{enumerate}
\setlength\itemsep{0.1em}
\item \href{https://en.wikipedia.org/}{Wikipedia} : the best-known free online encyclopedia, maintained by volunteers; everyone is allowed to edit pages, while moderators maintain the quality of the content.
\item \href{https://powo.science.kew.org/}{Plant of the World Online} (PoWO): an international collaborative database of the world's flora. The data are based on scientific publications and are maintained by the Royal Botanic Gardens of Kew (\cite{botanic}.
\item  \href{http://www.llifle.com/}{Encyclopedias of Living Forms} (LLifle): a collaborative effort to provide species descriptions, offering descriptions of 31,213 plant species, with a focus on xerophytes.
\item \href{https://apps.worldagroforestry.org/treedb/index.php}{World Agroforestry Center} (ICRAF): which provides textual descriptions of 670 tree species that are useful in agroforestry.
\end{enumerate}
%
%
All these sources contain structured plant species information divided into different sections, \emph{e.g.}, ``Introduction'', ``Appearance'', ``Characteristics'', and ``Habitat'', which are specific to each of source. These section headers allow for an automatic labelling process of the data; for example, text stemming from ``Introduction'' and ``Habitat'' are assigned non-descriptive labels, \emph{i.e.}, they are not relevant to the description of the species, while text from ``Characteristics'' and ``Appearance'' is assigned to the descriptive class. We also consider random pages from Wikipedia, not pertaining to species, as an augmentation approach that enriches the non-descriptive data.

\paragraph{\underline{\normalfont{Training the classifier.}}}
We can then proceed with training a description detector.
For this task, we need a model that is able to assign a binary label (description \emph{versus} non-description) to a piece of text of arbitrary length.
The most straight forward approach for this is to use a text encoder model, that can convert a text sequence of any size (up to some maximum allowed length) into a vector of fixed length. 
Any machine learning classifier can then be trained, in a supervised manner, using this vector representation as input.
For our description sentence classification model, we turn to a distilled version of BERT (Bidirectional Encoder Representation from Transformers)~\cite{devlin2018bert}, a widely used NLP model for obtaining vector representations of sentences, and specifically to DistillBERT (\cite{sanh_distilbert_2020}. This decision was motivated by the balance between complexity and performance that DistillBERT exhibits. This variant comprises $40\%$ less parameters than the original BERT model, leading to $60\%$ faster computations, while still yielding $97\%$ performance on general language understanding.
Both BERT and DistillBERT have been trained on a large corpus of English text, and pre-trained model weights are freely available. 
Within this context, we augment the base DistillBERT model by introducing: (i) a dropout layer for regularization purposes and (ii) two fully connected layers: the first layer takes the output vector of DistillBERT, of size $768$, and outputs a vector of size $512$, while the second layer takes this output vector and yields an output of size $2$, which are the logits for our binary classification task.

To prepare the collected text for fine-tuning the model, the first step is to split it into discrete tokens, which are either syllables or entire words, for which we use a tokenizer~\cite{wolf_huggingfaces_2020}. The conventional BERT architecture can accommodate up to $512$ tokens at once, corresponding to approximately 400 words.
This means that text spans longer than 512 tokens (\emph{e.g.}, a paragraph or a sentence) need to be truncated to length 512 in order to be compatible with DistillBERT.
In this work,  we randomly split the text into text spans with a minimum of $10$ and a maximum of $512$ tokens. 
This works as data augmentation and forces the model to capture the characteristics of descriptive sentences, even when not seeing the whole sentence, potentially providing robustness when exposed to text in the wild, where sentences could have different length and structure compared to the training set. 
We train and validate the model only on data stemming from Wikipedia and PoWO, while text from LLifle and ICRAF are used exclusively for model evaluation. By using different data sources for training and testing, we can obtain a better estimate of the generalization performance of the method on arbitrary websites returned by the search engine API upon deployment. For the evaluation, we use a sentencizer~\cite{honnibalspacy2020} to split the text into sentences.

\paragraph{\underline{\normalfont{Noise robust loss function.}}}
Due to the use of automatically obtained labels, it is likely that the resulting dataset will contain inconsistencies, as is common when dealing with unstructured data~\cite{kumar_text_2020}; after all, not all text within the description section of a Wikipedia article consists of descriptive sentences, and some descriptive sentences may occur outside of it, typically in the introductory sections. There also exists a chance that some section headers may be missed through this process, further increasing the amount of noise in the final dataset. 
We can mitigate the effects of this potential inconsistency by turning to classification losses that are designed for robustness against noisy labels.
We chose the ``soft bootstrap'' consistency objective~\cite{Reed:2015,marcos2022attribute}. The underlying principle is that the labels are ``diluted'' by the model's current prediction, thus reducing the impact on the loss of data points in which the model confidently disagrees with the label. Specifically, the loss is computed as:
\begin{equation}
\centering
    \mathrm{SoftLoss}(q,t) = \sum_{k=1}^L [ \beta t_k + (1-\beta) q_k]\log q_k
    \label{eqn:softloss}
\end{equation}
where $\mathbf{q}$ are the predicted class probabilities, $\mathbf{t}$ are the observed noisy labels and $\beta$ is a balancing factor between the current prediction and the target. In this way, we can use the current state of the model to dynamically adapt the prediction targets, allowing the model to pay less attention to inconsistent labels. As the model improves its predictions over time, it becomes more coherent, allowing for assessment of the consistency of the noisy labels. We set $\beta=0.20$ in a similar fashion to previous works (\cite{zhang_learn_2020,marcos2022attribute}.

For fine-tuning the model, we keep the DistillBERT parameters frozen and train only the added classification head. We use the Adam optimizer to minimize Eq. \eqref{eqn:softloss} with a learning rate of $3\cdot 10^{-5}$, a batch size of $32$ and gradient clipping with a norm of $1.0$. The model is fine-tuned until convergence, for a total of 35 epochs.

\subsection*{Trait information extraction}

\subsubsection*{Information extraction with a generative LLM}

The next step towards information extraction for species descriptions involves extracting relevant information from the obtained text snippets into a structured form. 
To this end, we leverage the recent advances in LLMs, which have been empirically demonstrated to capture relational knowledge in the training data that can be extracted via natural language queries, also known as prompts~\cite{ouyang2022training}. 
These models have been shown to perform well on relatively generic tasks, such as common sense knowledge~\cite{davison-etal-2019-commonsense} or general knowledge~\cite{petroni-etal-2019-language}. 

Although it is possible to directly query an LLM with a question, without providing any additional information, one should be aware about their tendency of providing responses that look legitimate, but that are completely unfounded, known as hallucinations~\cite{zhang2023siren}. 
This is even more prominent in specialized domains, including botany, that are characterized by long-tailed distributions in which a few elements are very abundant and many are extremely rare. 
This leads to LLMs being unreliable for this majority of uncommon elements. 
To mitigate this issue, we turn the task into information extraction from text via search engine retrieval~\cite{lewis2020retrieval}. We achieve this by feeding the LLM a piece of descriptive text, obtained via the  harvesting approach detailed in the \textit{``Textual data harvesting and description detection''} section, along with questions referring to a predetermined set of traits and possible trait values.
At the same time, we give the LLM the option to explicitly state if the requested information is not available (NA) in the given text, mitigating potential hallucination issues that could otherwise arise.
This means that only a subset of traits will be assigned a trait value, the proportion of which we will refer to as \emph{coverage rate}.

\subsubsection*{Choice of LLM}

The 32k-token context window in mistral-medium, by Mistral AI, was sufficient to accommodate all the text for a given species along with the considered traits and trait values.
In addition to this, we found mistral-medium (version 2312, released on December 2023) to be a good compromise in preliminary tests, with results substantially better than those of GPT-3.5 Turbo and at a similar cost.
We tested the Mixtral-8x7B and LLaMA 2 open source models, but they provided results of lower quality than those of ChatGTP-3.5 Turbo.
However, we have conducted tests with the open source Mixtral-8x22B, which was released after we conducted most of our experiments, and obtained results comparable to mistral-medium, making it a good alternative for labs with access to a GPU cluster. Refer to Section ``Additional experimental results'' for details.

\subsubsection*{Prompt design}

The considered species-trait datasets comprise categorical traits; these can be encoded in a binary form and expressed as multiple choice textual questions to engineer discrete prompts. In this binary encoding context, we are interested in discovering which trait values should be ``1'' or ``0'' for any given set of species and trait value by exploiting the information from the retrieved description sentences. In Table~\ref{tab:species_binary}, such an encoding of the categorical traits ``Life form'' and ``Phyllotaxis'' of the Caribbean dataset is depicted.
\begin{table}[h]
    \centering
    \begin{tabular}{l|ccccc}
         \multirow{2}{*}{\textbf{Species}} &  \multicolumn{2}{c}{\textbf{Life form}} & \multicolumn{2}{c}{\textbf{Phyllotaxis}} & $\cdots$\\
          & \textit{Tree} & \textit{Liana} & \textit{Alternate} & \textit{Opposite} & $\cdots$ \\\hline
         Avicennia germinans & 1 & 0 & 0 & 1 & $\cdots$\\
         Metopium brownei & 1 & 0 & 1 & 0 & $\cdots$\\
         Cynophalla flexuosa & 0 & 1 & 1 & 0 & $\cdots$ \\\hline
    \end{tabular}
    \caption[Binary encoding example of the manual annotations.]{A binary encoding of the manual annotations of two different traits for three different species. The entries denote the presence (``1'') or absence (``0'') of a particular trait for each species.}
    \label{tab:species_binary}
\end{table}
Thus, for each trait, we first group together all its possible values, \emph{e.g.}, ``Life form'': [``tree'', ``liana''] and ``Phyllotaxis'': [``alternate'', ``opposite'']. 
Then, to create a prompt for the  LLM, we consider: 
(i) all the textual description sentences about a species, and 
(ii) the list of traits and considered trait values as described before. 
Based on this construction, we prompt the LLM to infer the values for each trait based on the provided text. 
A realistic example prompt based on the described process is depicted on Figure~\ref{fig:prompt} (left). 
In this example, we ask the LLM about three traits. For the first two, ``Plant type'' and ``Phyllotaxis'', there is some information available in the input text: ``[...] is a deciduous tree'' and ``Leaves are alternate''. For the third trait, related to ``Trunk and root'', no information is present in the text.
Indeed, the actual response of the LLM when queried with the constructed prompt is shown on Figure~\ref{fig:prompt} (right). Therein, we observe that the LLM correctly infers the values of each trait from the given textual description, exhibiting behavior consistent to what we expected. 
For the last trait there is no evidence for any of the accepted trait values, and this will be treated as an NA.

\begin{figure}
\centering
\fbox{
\begin{minipage}{0.67\textwidth}
\begin{footnotesize}

\texttt{ \scriptsize
We are interested in obtaining botanical trait information about the species Albizia coriaria.
\\
\\
We will provide an input text with botanical descriptions, followed by a dictionary where each key 'name' represents a trait name, referring to specific organ or other element of the plant, and is associated to a list with all possible trait values for that trait, ['value\_1', 'value\_2', ..., 'value\_n'].
\\
\\
Input text:
\\
Albizia coriaria is a deciduous tree 6-36 m tall.
The flowers are subsessile or on pedicels 0.5-2 mm long, minute bracteoles, 1.5-2 mm long usually falling off before flowering.
The fruit is an oblong, flat pod, densely but finely pubescent, transversely veined, and pale brown when ripe.
Medium-size tree.
Leaves are alternate, bipinnately compound with 3 to 10 pairs of pinnae with ovate to lanceolate stipules and leaflets in 5 to 17 pairs per pinna.
It is heavily branched forming a spreading dome-shaped crown.
\\
\\
Initial dictionary of traits with all possible values:
\\
\{"Plant type": ['Tree', 'Shrub', 'Bush', 'Ficus strangler', 'Liana', 'Parasitic', 'Palm tree', 'Herbaceous'], 
\\
"Phyllotaxis": ['Phyllotaxis alternate', 'Opposite phyllotaxis', 'Phyllotaxis whorled'],
\\
"Trunk and root": ['Base of trunk straight', 'Base of trunk flared', 'Foothills', 'Stilt roots', 'Aerial roots']\}
\\
\\
Turn each string s in the list of values in the dictionary into a sublist (s,b), where b is a binary number,either 0 or 1, indicating whether there is strong evidence for value s in the input text. Double check that 'value\_i' is reported referring to trait 'name' in the text, and not to a different trait. Always set 'b' to '0' if you are not 100\% sure about the association. Do not add new trait values and do not modify the initial ones.Return the dictionary of traits and sublists of (value, evidence) containing all possible names and (values, evidence) tuples.
\\
\\
Output only a dictionary in JSON format, no other text at all.
}
\end{footnotesize}
\end{minipage}}
\hfill
\centering
\fbox{
\begin{minipage}{0.27\textwidth}
\begin{footnotesize}
\texttt{ \scriptsize
\{\\
"Plant type": [
\\
("Tree", 1),\\
("Shrub", 0),\\
("Bush", 0),\\
("Ficus strangler", 0),\\
("Liana", 0),\\
("Parasitic", 0),\\
("Palm tree", 0),\\
("Herbaceous", 0)],\\
"Phyllotaxis": [\\
("Phyllotaxis alternate", 1),\\
("Opposite phyllotaxis", 0),\\
("Phyllotaxis whorled", 0)],\\
"Trunk and root": [\\
("Base of trunk straight", 0),\\
("Base of trunk flared", 0),\\
("Foothills", 0),\\
("Stilt roots", 0),\\
("Aerial roots", 0)]\\
\}
}
\end{footnotesize}
\end{minipage}}
\caption[An example prompt querying the LLM about morphological traits presence.]{An example prompt used to query the LLM about the presence of morphological traits given a textual description sentence for a given species (left), along with the LLM response (right). The LLM correctly identifies that there is evidence in the text indicating that the plant type is tree and the phyllotaxis, alternate, while no evidence can be found for the other trait values.}
\label{fig:prompt}
\end{figure}

Although the example prompt includes only three traits and provides information about a single species, we are interested in scaling this approach to hundreds/thousands of species and long lists of possible traits.
Scaling the approach can be done by simply repeating the process for new species and including additional traits in the prompt. In our work, and to be able to compare the prompt results to the ground truth data of the three considered datasets (Caribbean, W. Africa and Palms), we consider the exact same species, traits, and trait values found in each dataset. 

By using an LLM with a large enough context window, it is possible to fit the whole text and dictionary of traits into a single prompt.
Alternatively, it is also possible to split the task into multiple prompts by querying about a single prompt at a time or by providing only a subset of the input text.
The answers of the LLM can then be parsed in order to build a species-trait matrix in the same format and the manual annotations.

\subsubsection*{Evaluation metrics}

\paragraph{\underline{\normalfont{Evaluation of the automatic trait extraction.}}}
In order to evaluate the responses of the LLM, we compare them to the species-trait matrices manually curated by expert botanists  (\emph{Caribbean}, \emph{Palms} and \emph{West Africa} datasets, described above).
We report the proportion of traits for which a value was found, i.e., the coverage, along with the precision, recall and F$_1$ score computed for the found traits.
Precision, recall and F$_1$ are computed only for traits with coverage.
The precision is the proportion of predicted positives, all the trait values predicted by the model as being present, that turns out to be correct according to the manual dataset.
The recall is the proportion of positives in the manual dataset that are retrieved by the approach.
In order to combine these two complementary metrics, the $F_1$ score consists of the geometrical average of precision and recall.

\paragraph{\underline{\normalfont{Evaluation of the false negative rate.}}}
Even though the described evaluation process allows for assessing whether the detected traits are correct, according to manually created species-trait matrices, it does not allow for quantifying the false negative rate of the LLM, \emph{i.e.}, whether all traits described in the text are \textit{effectively} extracted. In this context, we need to assess whether the false negatives arise due to the LLM extraction process or due to the fact the information is simply not present in the harvested text. 
To mitigate this issue, we perform an additional evaluation of the trait extraction process by asking three senior botanists whether a certain trait value can be inferred by using a specific piece of text and using this information as ground truth.
Specifically, we first randomly selected a trait from one of the species-trait datasets. 
We then selected a random species from the same dataset and picked a text snippet with a low distance in the DistillBERT embedding space to the name of the trait, in order to increase the number of relevant text-trait pairs.
This allowed us to generate 1216 text-trait pairs which we then shared with the botanists. They were given the text and the ground truth trait value and asked if this trait value could be inferred from the text. 
Four senior botanists contributed to these responses.
According to the botanists, 298 out of the 1216 snippets contained relevant information about the trait of interest. 
To assess the capacity of the LLM extraction process in this setting, we construct the corresponding prompts with the same pairs of sentences and traits as the ones presented to the botanists. The prompt used was of the same structure as the one shown in Figure~\ref{fig:prompt}.
This allows us to investigate whether the LLM behaves in an excessively conservative manner, preferring to return empty results rather than make a mistake, or has a tendency to hallucinate responses that are not explicitly in the text.

\section*{RESULTS}

\begin{table}[htpb]
\centering
\begin{tabular}{@{}l|cc|cc|cc|cc@{}}
\cmidrule(l){2-9}
 & \multicolumn{2}{c}{\textbf{Precision}} & \multicolumn{2}{c}{\textbf{Recall}} & \multicolumn{2}{c}{\textbf{F1-score}} & \multicolumn{2}{c}{\textbf{\# sentences}} \\ \cmidrule(l){2-9} 
                 & Val. & Test & Val. & Test & Val & Test & Val    & Test \\ \midrule
Non-descriptive            & 0.98 & 0.94   & 0.99 & 0.98   & 0.99 & 0.96   & 167,955 & 66,246 \\
Descriptive      & 0.97 & 0.79   & 0.95 & 0.55   & 0.96 & 0.65   & 57,864  & 8,590  \\ \midrule
\end{tabular}
\caption[Precision-recall metrics test and two left out datasets]{The precision-recall metrics for the binary classification model tested on the test dataset and two external datasets (LLifle and AgroForestry).}
\label{tab:description_detection_metrics}
\end{table}

\begin{figure}[h]
    \centering

    \begin{tabular}{cc}
    \begin{subfigure}{0.85\textwidth}
        \framebox{\parbox{\dimexpr\linewidth-2\fboxsep-2\fboxrule}{%
        "
        \sethlcolor{first_color}\hl{Hedera helix is an evergreen climbing plant. < 0.586 >}
        \sethlcolor{second_color}\hl{The leaves are alternate. < 0.819 >}
        \sethlcolor{third_color}\hl{The house is large with enormous windows. < 0.045 >}
        \sethlcolor{fourth_color}\hl{This is something random, but the flowers are individually small. < 0.292 >}
        \sethlcolor{fifth_color}\hl{Metropolitan France was settled during the Iron Age by Celtic tribes. < 0.015 >}
        \sethlcolor{sixth_color}\hl{The fruit are purple-black to orange-yellow berries. < 0.930 >}
        \sethlcolor{seventh_color}\hl{One to five seeds are in each berry. < 0.708 >}
        \sethlcolor{eighth_color}\hl{Hedera helix prefers non-reflective, darker and rough surfaces with near-neutral pH. < 0.193 >}
        \sethlcolor{ninth_color}\hl{The petiole is 15-20 mm. < 0.824 >}
        \sethlcolor{tenth_color}\hl{Hedera helix is a species of flowering plant of the ivy genus. < 0.345 >}
        \sethlcolor{eleventh_color}\hl{Once Hedera Helix is established it is very difficult to control or eradicate. < 0.106 >}"}}
        \end{subfigure} &
        
        \begin{subfigure}[b]{0.065\textwidth}
        \raisebox{\dimexpr-0.5\height}{
            \includegraphics[width=1 \textwidth]{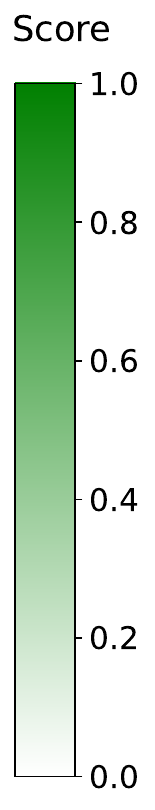} 
        }
        
        \end{subfigure}
        \end{tabular}

    \caption[An example set of sentences and their corresponding ``description'' scores.]{An example set of sentences and their corresponding ``description'' score. The text is broken into single sentences by the sentencizer of \cite{honnibalspacy2020} and the classifier classifies each sentences. Sentences with a value of 0.50 or higher are stored in the database. The darker the colour green, the higher the prediction value. The prediction value is also shown after each sentence.}
    \label{fig:webcrawler_sents}
\end{figure}

\subsection*{Descriptive text classification}

The descriptive/non-descriptive dataset creation process described in the \textit{``Descriptive text classification''} section, resulted in approximately $1.45$ million sentences; $1.1$ million sentences corresponding to non-descriptive text and $356k$ corresponding to descriptive text. 
In the performance analysis of the model, shown in Table~\ref{tab:description_detection_metrics}, we observe that, within the in-domain validations set, our description classification model reaches very high precision for both classes, \emph{i.e.}, ``Description'' and ``Non-Description'', with F$_1$ scores of 0.96 and 0.99 respectively. However, the recall in the test set drops substantially for the descriptive class, from 0.95 to 0.55, albeit not for the non-descriptive class, which is 0.98 in the test set. 
A few example sentences with their corresponding score can be seen in Figure~\ref{fig:webcrawler_sents}, where we can see that the model behaves as expected, with only botanical descriptions having a score higher than 0.5.

\subsection*{Descriptive text harvesting}

Having trained and validated our descriptive sentence detector, we can now consider any potential source of textual information to extract species descriptions towards a downstream task. 
The text harvesting step returned description text for the majority of species, but not for all. Specifically, we obtained text for 40/42 species in the Caribbean dataset, 358/361 for the West Africa dataset and 248/333 for Palms. On average, we obtained 35, 36.8 and 43.5 descriptive sentences per species for each dataset respectively. Refer to the code repository to find the whole set of found descriptive sentences and their original URL (\url{https://github.com/konpanousis/AutomaticTraitExtraction/blob/main/Descriptions}.

\subsection*{Automatic trait extraction}

\subsubsection*{Comparison to manually curated trait data}

\begin{table}[htb]
    \centering
    \begin{tabular}{l|c|c|c||c}
        \hline
         
          Dataset  & Precision & Recall & F$_1$ & Coverage\\\hline
          Caribbean& 0.7493 & 0.7800 & 0.7643 & 0.5500\\ 
          West Africa & 0.8058 & 0.7776 & 0.7806 & 0.5588\\
          Palms & 0.7013 & 0.7706 & 0.7343 & 0.5584
          \\\hline
          
    \end{tabular}
    \caption[Trait prediction precision, recall, F$_1$ score and coverage.]{Precision, recall and F$_1$ score with respect to the three manually curated databases, along with the coverage, \emph{i.e.}, the proportion of trait-species entries for which at least one value is found. The accuracy metrics are computed only for these entries.}
    \label{tab:traits_accuracy}
\end{table}

As shown in Table~\ref{tab:traits_accuracy}, the coverage ranges between 55\% and 56\%, meaning that over half of traits are assigned a value with the described method.
The F$_1$ scores range between 73\%, in the Palms dataset, and 78\% in the West Africa dataset, with the recall being remarkably constant, between 77\% and 78\%, and the precision varying between 70\% in Palms and 80\% in West Africa.

\begin{figure}[htb]
     \centering
     \begin{subfigure}[b]{0.8\textwidth}
         \centering
         \includegraphics[width=\textwidth]{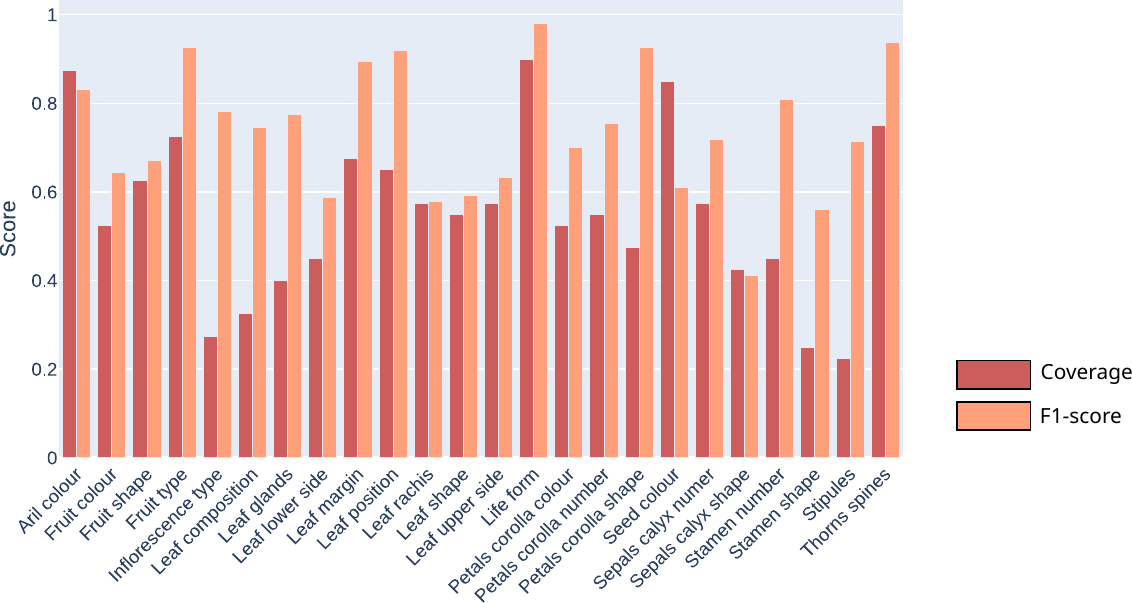}
         \caption{\emph{Caribbean} dataset.}
         \label{fig:caribbean}
     \end{subfigure}
     \hfill
     \begin{subfigure}[b]{0.6\textwidth}
         \centering
         \includegraphics[width=\textwidth]{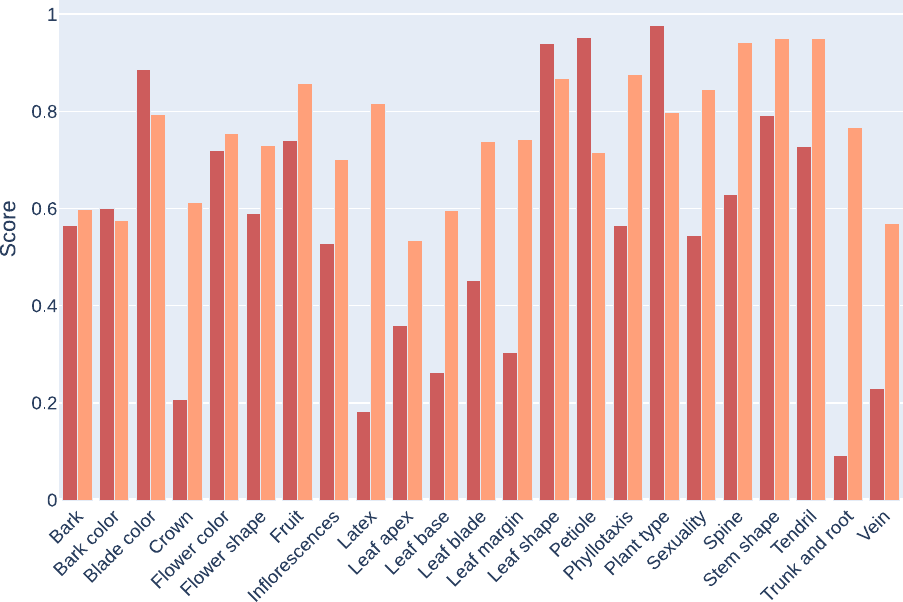}
         \caption{\emph{West Africa} dataset.}
         \label{fig:plantnet}
     \end{subfigure}
     \hfill
     \begin{subfigure}[b]{0.39\textwidth}
         \centering
         \includegraphics[width=\textwidth]{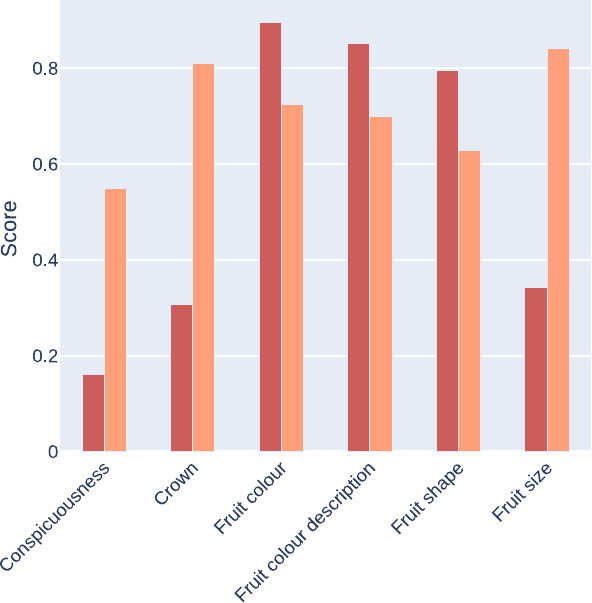}
         \caption{\emph{Palms} dataset.}
         \label{fig:palms}
     \end{subfigure}
        \caption[Coverage and F$_1$-score per dataset and trait type.]{F$_1$-score (orange) and coverage (red) per trait with respect to the three manually curated databases. The coverage is the proportion of species for which at least one value is found. The F$_1$-score is computed only for these species.}
        \label{fig:coverage_f1}
\end{figure}

The per-trait F$_1$ scores and coverages are displayed in Figure~\ref{fig:coverage_f1}. Here we see large variations in both aspects. Some commonly found traits, such as \textit{life form} and \textit{seed color} in the Caribbean dataset or \textit{plant type} and \textit{leaf shape} in the West African dataset, have been retrieved for well above 80\% of species, while \textit{trunk and root} in the latter are only found for around 10\% of species.
Also, large variations in terms of F$_1$ accuracy can be observed across traits in all datasets. We see a tendency towards higher accuracy in traits for which fewer values are allowed. For instance, \textit{life form} in Caribbean has only two possible values, and the F1 stands at over 95\%. On the other hand, \textit{fruit color} in Palms, has around 70\% F$_1$ for 12 possible values.

\begin{figure}[h!]
     \centering
     \begin{subfigure}[b]{0.49\textwidth}
         \centering
         \includegraphics[width=\textwidth]{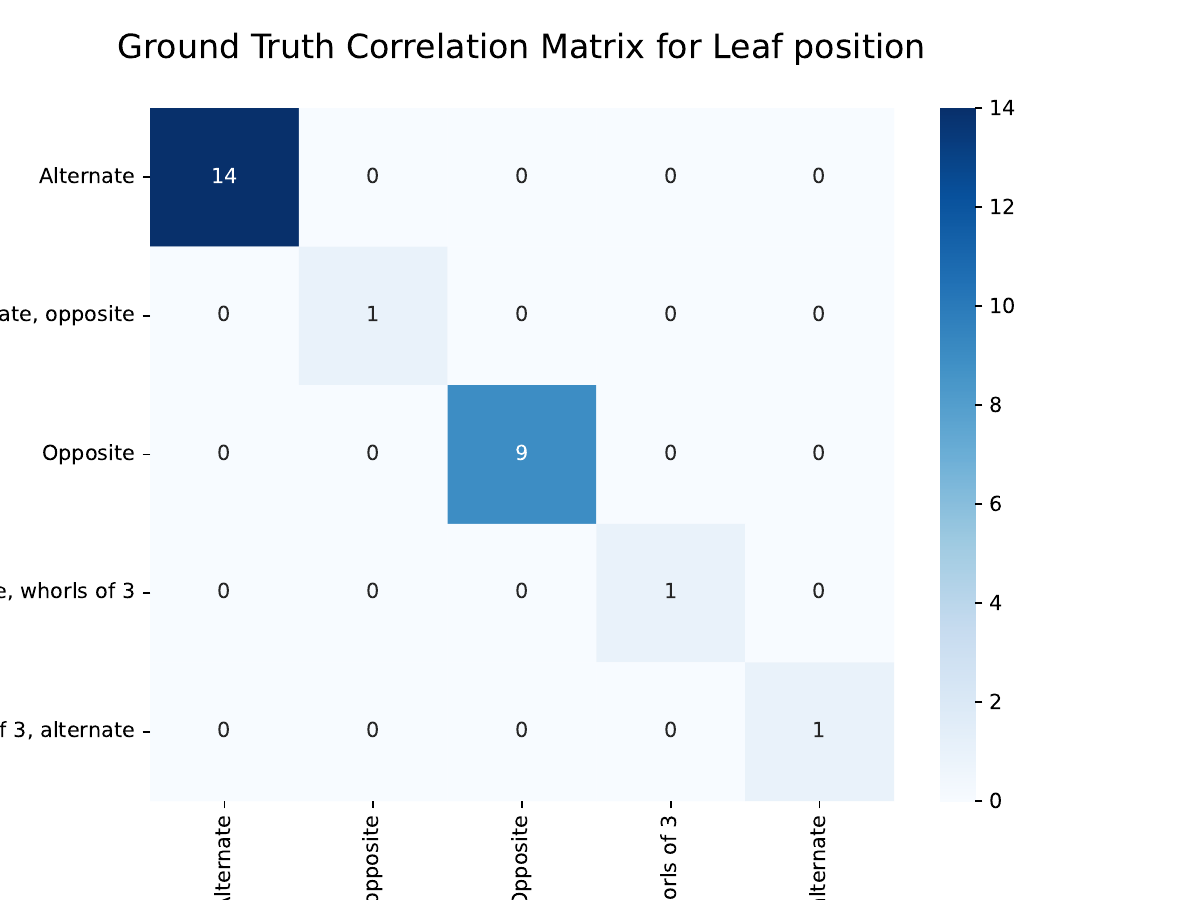}
         \caption{Leaf position (Caribbean dataset) within the annotations.}
         \label{fig:Caribbean_Leaf position_true}
     \end{subfigure}
     \begin{subfigure}[b]{0.49\textwidth}
         \centering
         \includegraphics[width=\textwidth]{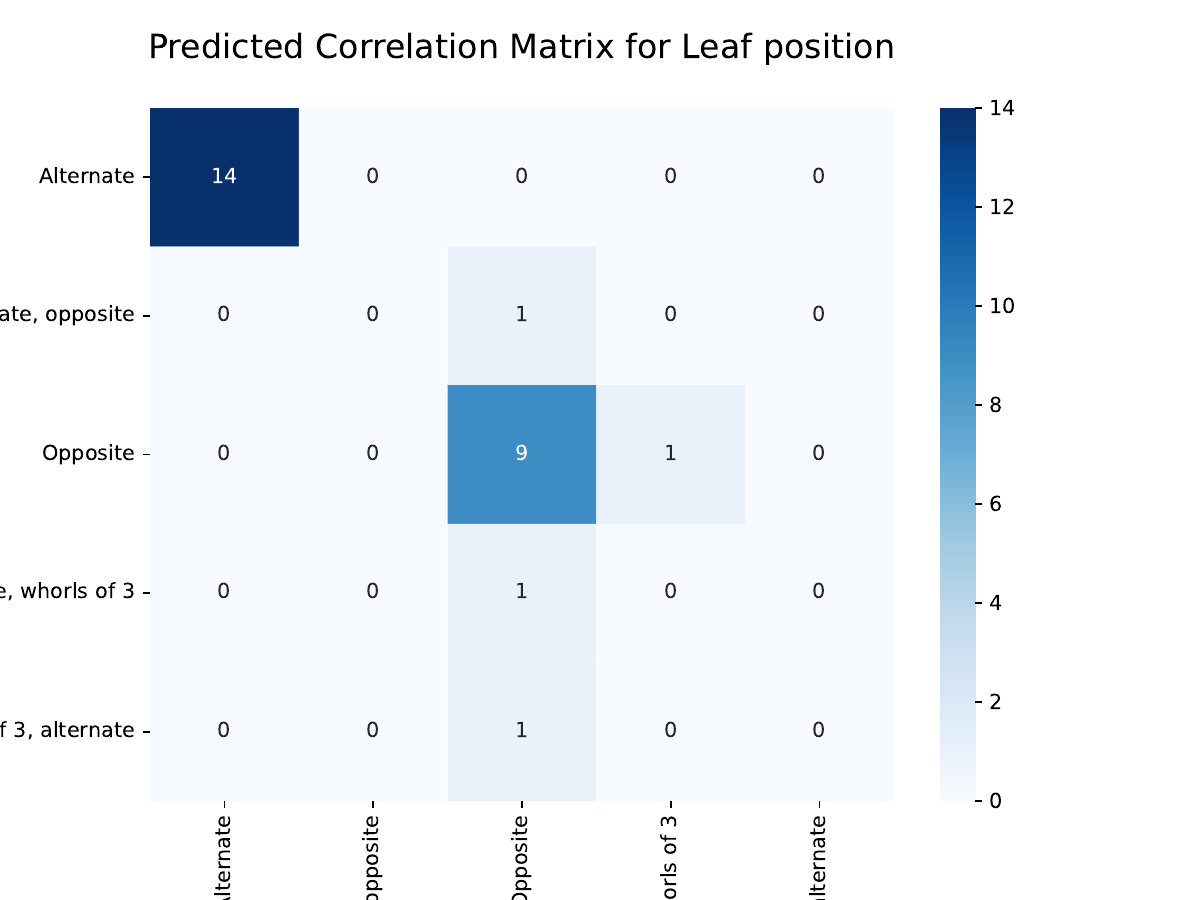}
         \caption{Leaf position (Caribbean dataset) between predictions (columns) and annotations (rows).}
         \label{fig:Caribbean_Leaf_position_pred}
     \end{subfigure}
     \begin{subfigure}[b]{0.49\textwidth}
         \centering
         \includegraphics[width=\textwidth]{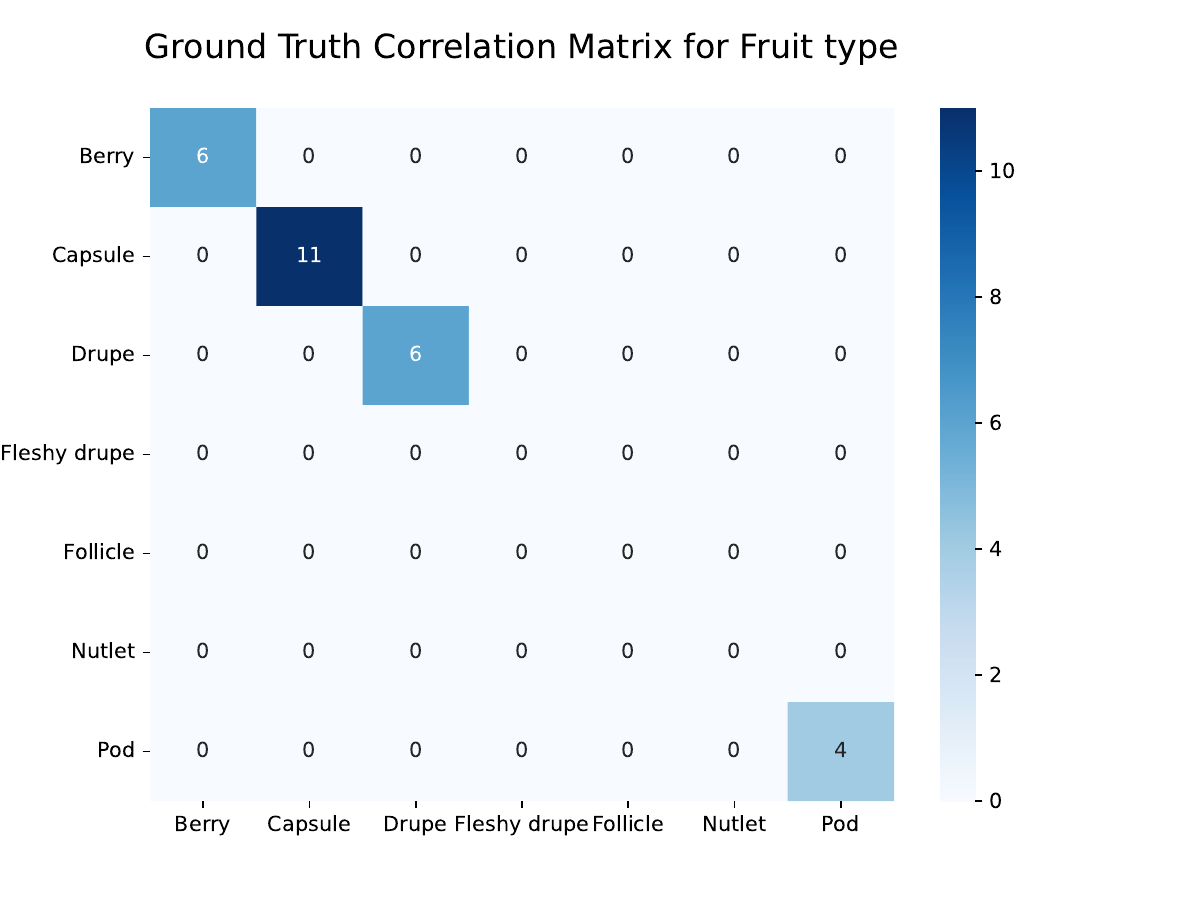}
         \caption{Fruit type (Caribbean dataset) within the annotations.}
         \label{fig:Caribbean_Fruit_type_true}
     \end{subfigure}
     \begin{subfigure}[b]{0.49\textwidth}
         \centering
         \includegraphics[width=\textwidth]{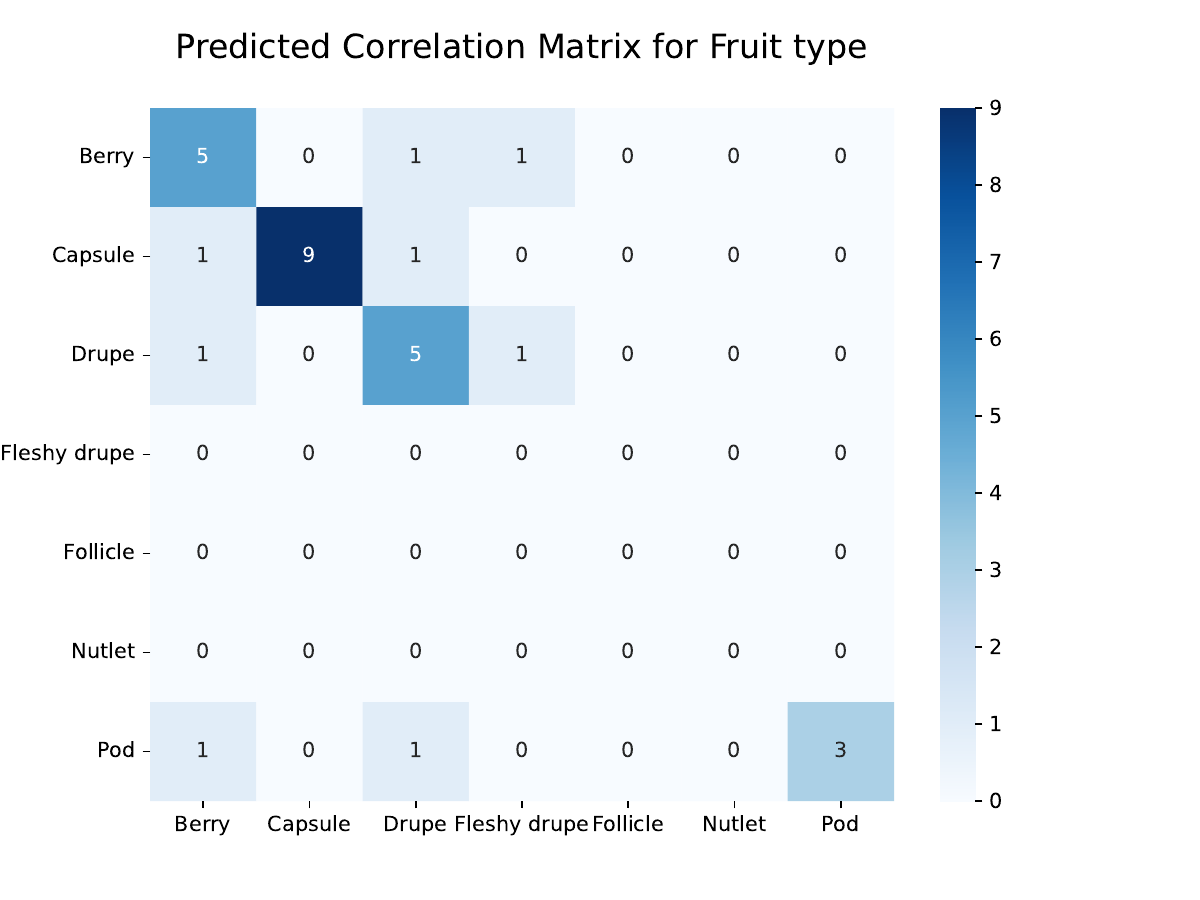}
         \caption{Fruit type (Caribbean dataset) between predictions (columns) and annotations (rows).}
         \label{fig:Caribbean_Fruit_type_pred}
     \end{subfigure}
     \begin{subfigure}[b]{0.49\textwidth}
         \centering
         \includegraphics[width=\textwidth]{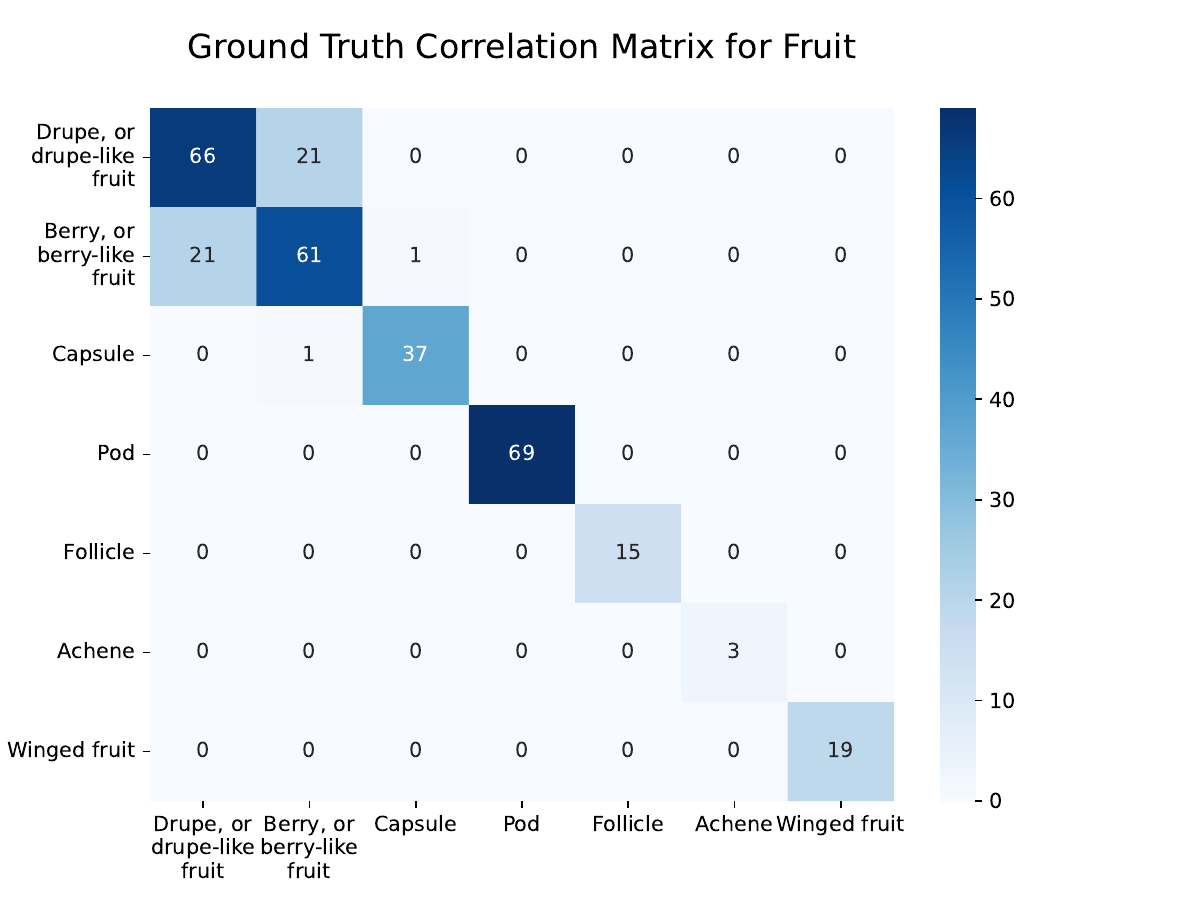}
         \caption{Fruit (W. African dataset) within the annotations.}
         \label{fig:Plantnet_Fruit_true}
     \end{subfigure}
     \begin{subfigure}[b]{0.49\textwidth}
         \centering
         \includegraphics[width=\textwidth]{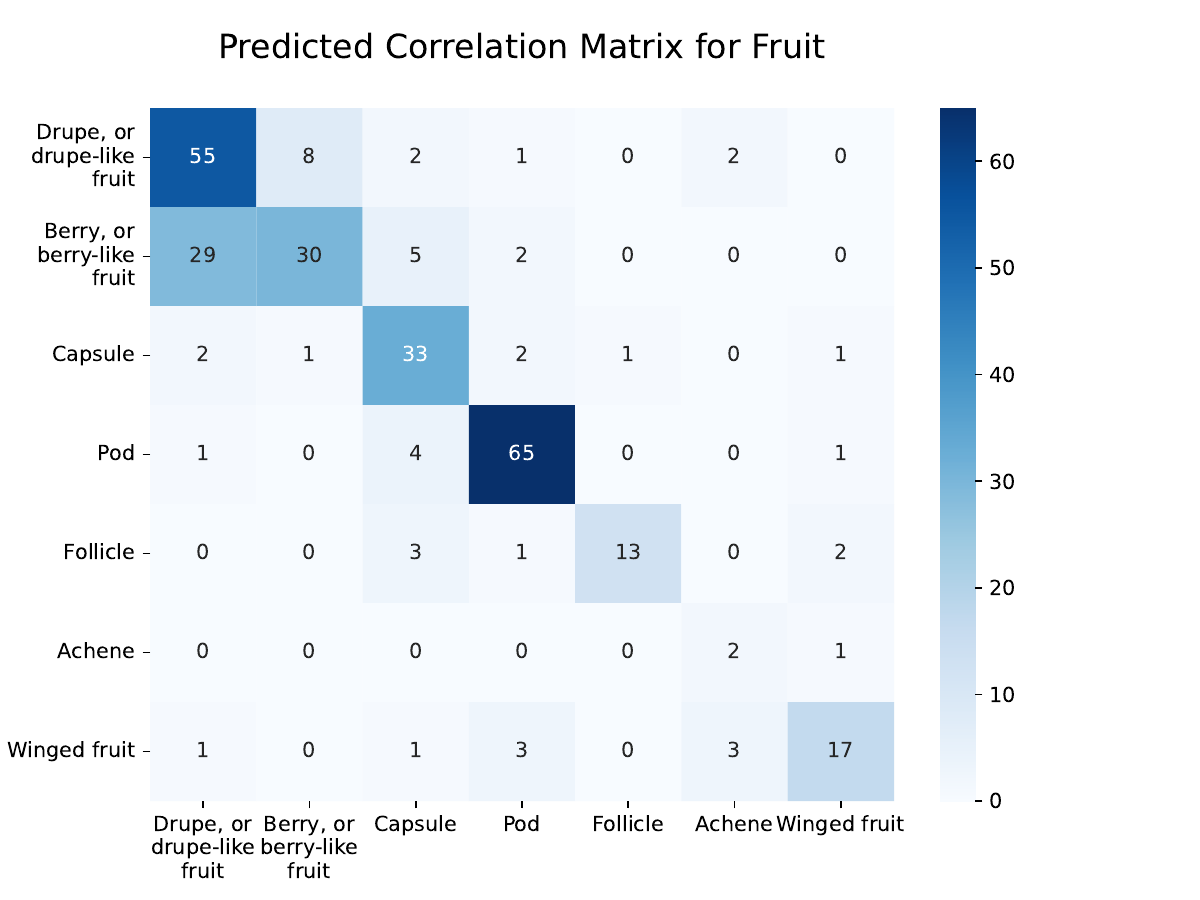}
         \caption{Fruit (W. African dataset) between predictions (columns) and annotations (rows).}
         \label{fig:Plantnet_Fruit_pred}
     \end{subfigure}
    
    \caption[Example co-occurrence matrices for \textit{leaf position}, \textit{fruit type} and \textit{fruit}.]{Co-occurrence matrices for every pair of trait values in \textit{leaf position} and \textit{fruit type} in the Caribbean dataset, and \textit{fruit} in the West Africa dataset (left), and the corresponding co-occurrences between the prediction and the annotations (right). We can see that the patterns of co-occurrence is maintained.}
    \label{fig:multi_conf_1}
\end{figure}

\begin{figure}[]
     \centering

     \begin{subfigure}[b]{0.49\textwidth}
         \centering
         \includegraphics[width=\textwidth]{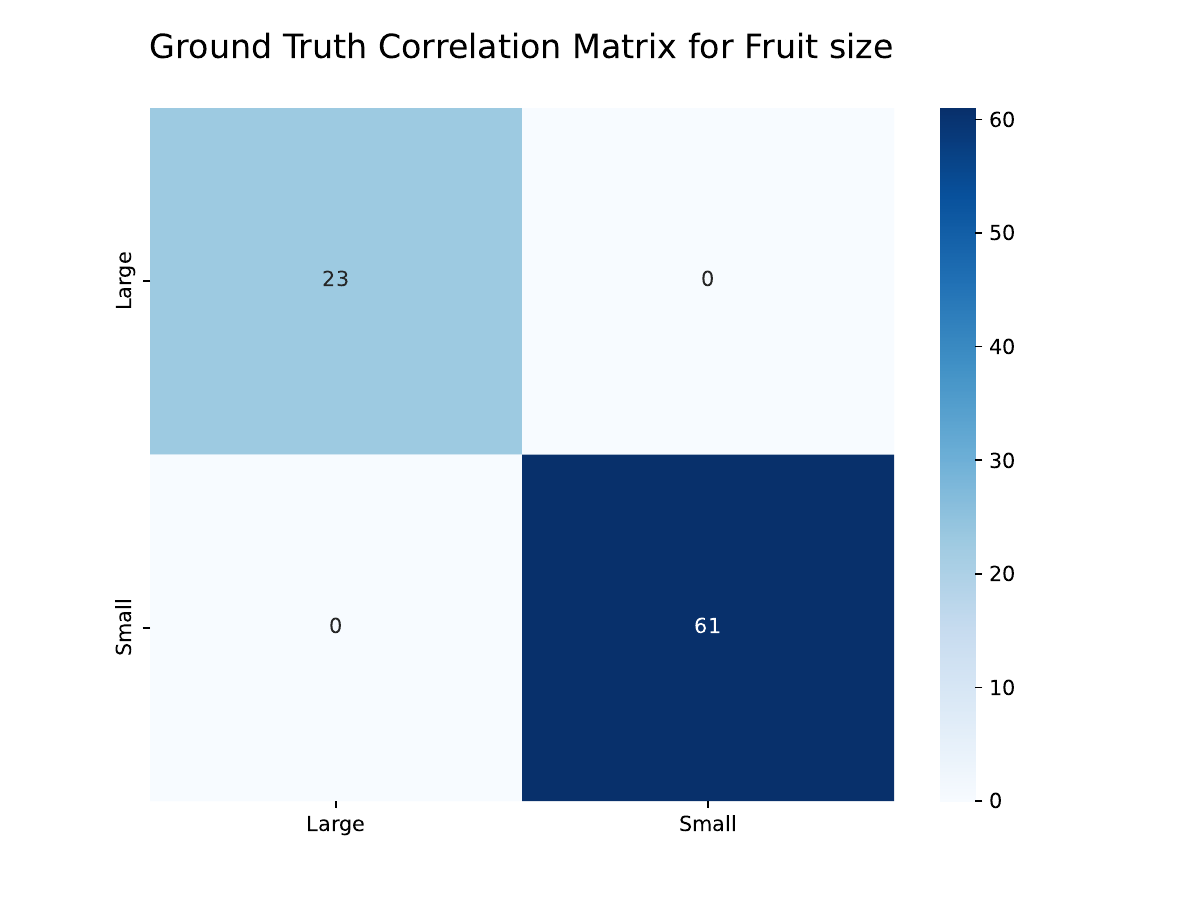}
         \caption{Fruit size (Palm dataset) within the annotations.}
         \label{fig:Palm_Fruit_size_true}
     \end{subfigure}
     \begin{subfigure}[b]{0.49\textwidth}
         \centering
         \includegraphics[width=\textwidth]{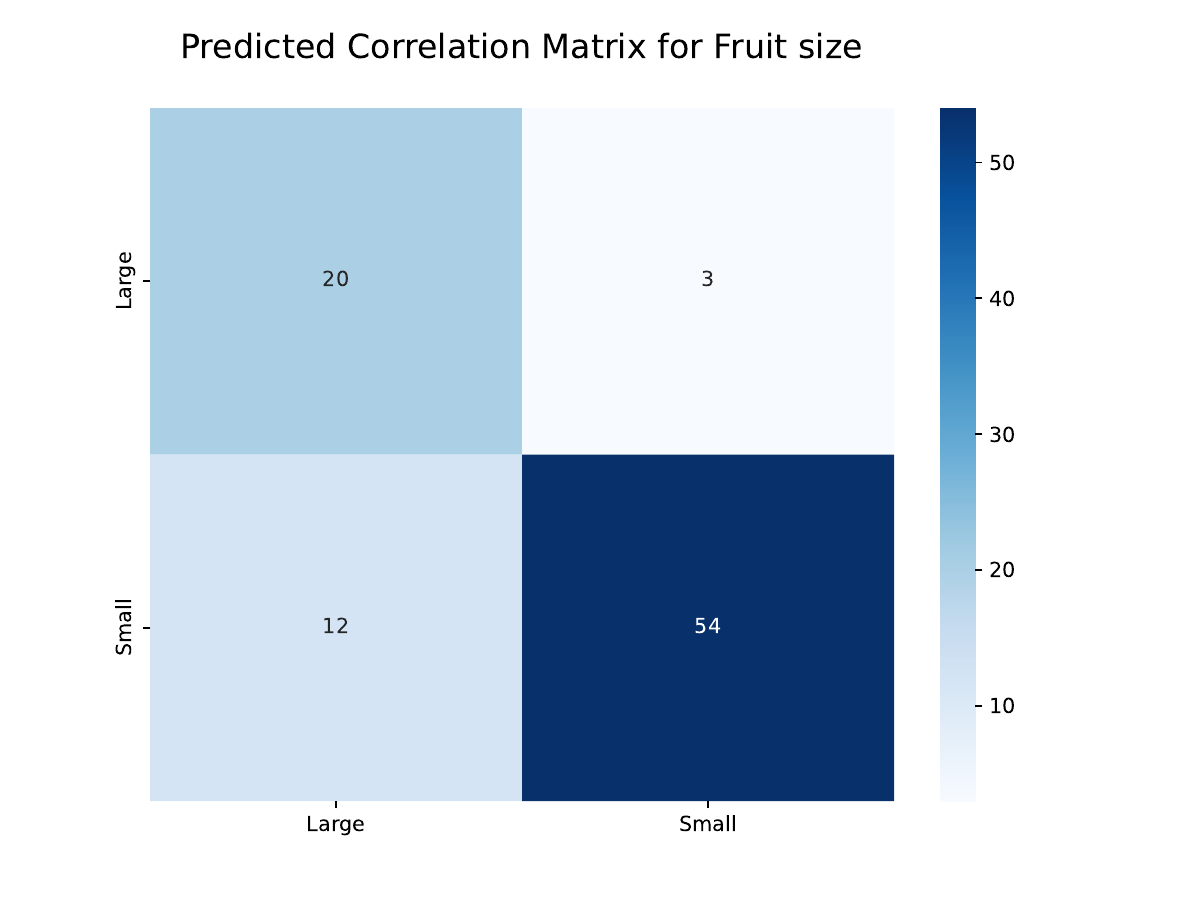}
         \caption{Fruit size (Palm dataset) between predictions (columns) and annotations (rows).}
         \label{fig:Palm_Fruit_size_pred}
     \end{subfigure}
     \begin{subfigure}[b]{0.49\textwidth}
         \centering
         \includegraphics[width=\textwidth]{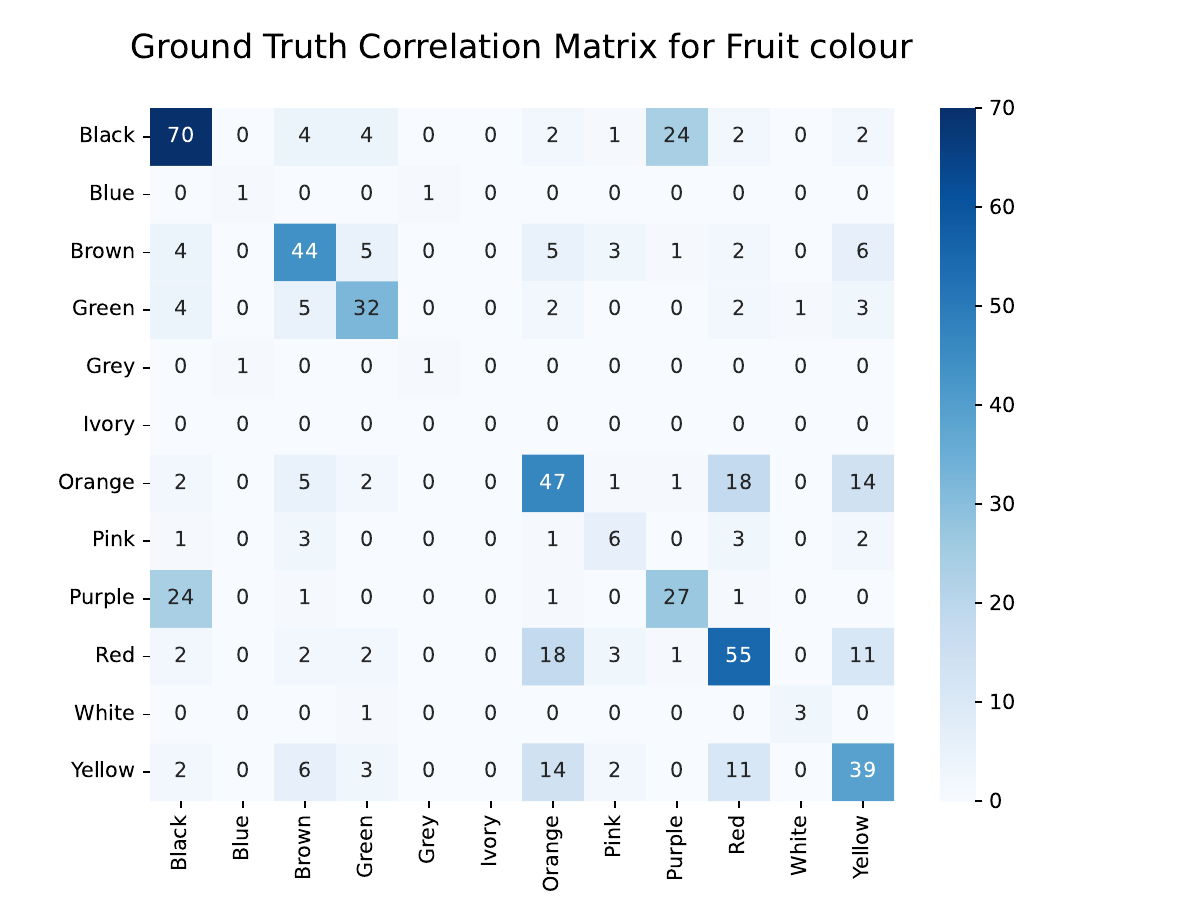}
         \caption{Fruit colour (Palm dataset) within the annotations.}
         \label{fig:Palm_Fruit_color_true}
     \end{subfigure}
     \begin{subfigure}[b]{0.49\textwidth}
         \centering
         \includegraphics[width=\textwidth]{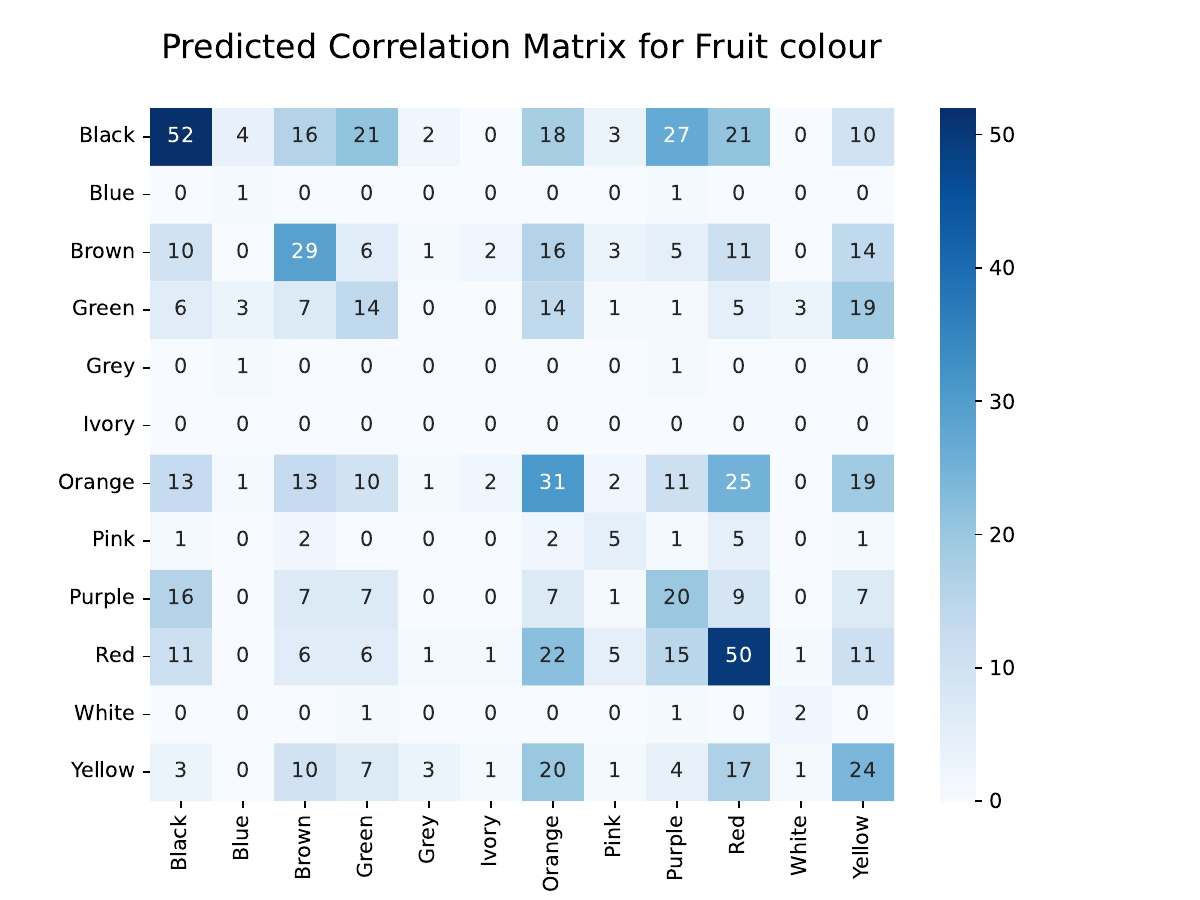}
         \caption{Fruit colour (Palm dataset) between predictions (columns) and annotations (rows).}
         \label{fig:Palm_Fruit_color_pred}
     \end{subfigure}
        \caption[Example co-occurrence matrices for some traits in all  datasets.]{Co-occurrence matrices for some traits in the three datasets. For each trait, we compare the co-occurrences between the annotations (left) and the co-occurrences between the predictions and the annotated values (right). We can see that the patterns of co-occurrence is maintained.}
        \label{fig:multi_conf_2}
\end{figure}

\begin{figure}[]
     \centering
     \begin{subfigure}[b]{0.49\textwidth}
         \centering
         \includegraphics[width=\textwidth]{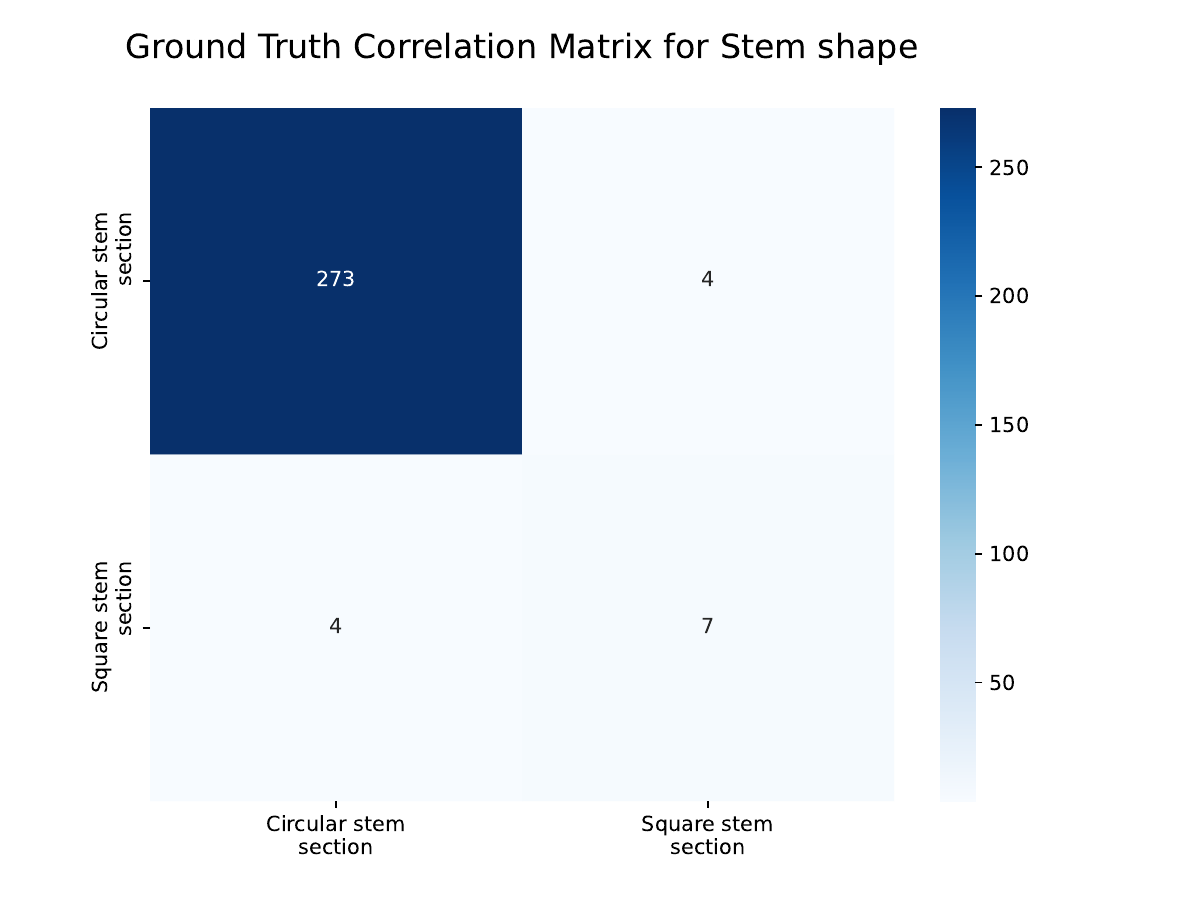}
         \caption{Stem shape (W. African dataset) within the annotations.}
         \label{fig:Plantnet_Stem_shape_true}
     \end{subfigure}
     \begin{subfigure}[b]{0.49\textwidth}
         \centering
         \includegraphics[width=\textwidth]{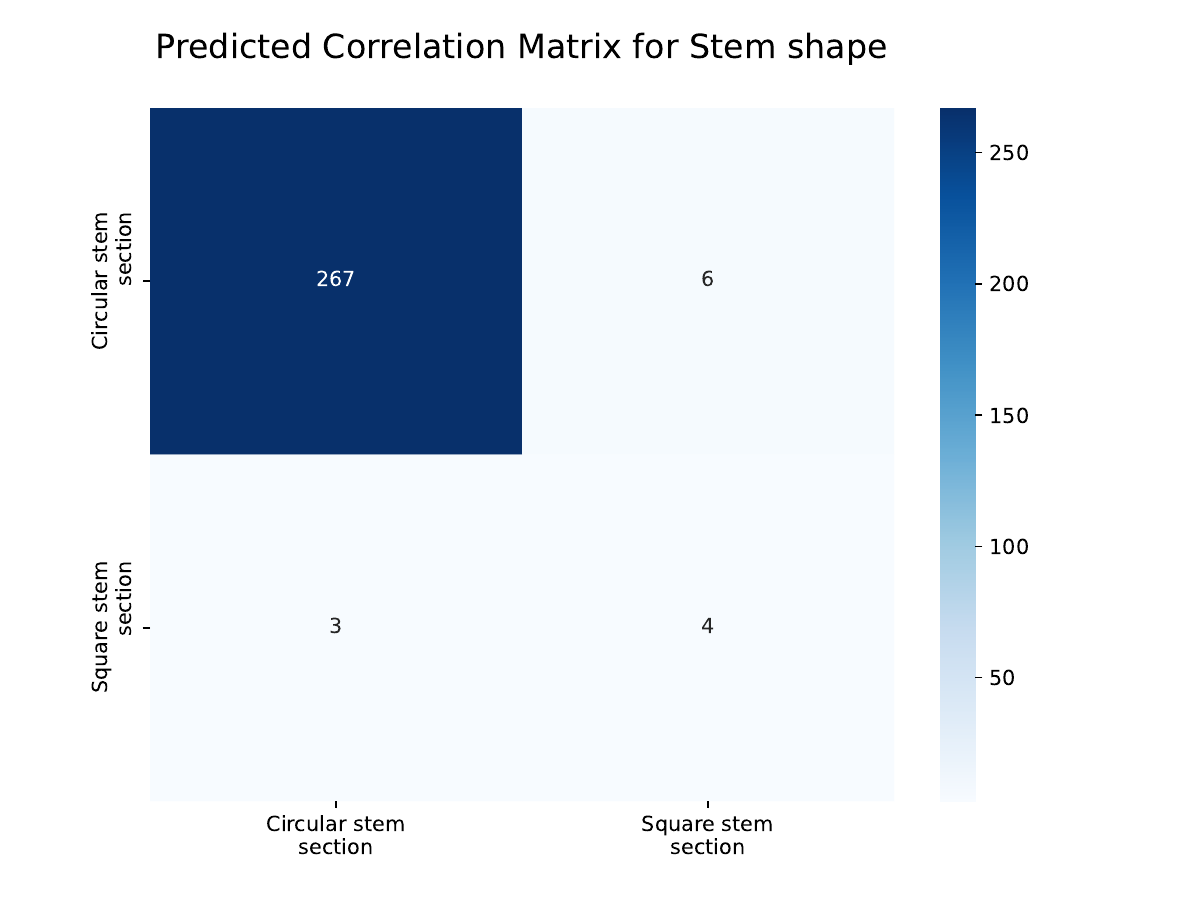}
         \caption{Stem shape (W. African dataset) between predictions (columns) and annotations (rows).}
         \label{fig:Plantnet_Stem_shape_pred}
     \end{subfigure}
     \begin{subfigure}[b]{0.49\textwidth}
         \centering
         \includegraphics[width=\textwidth]{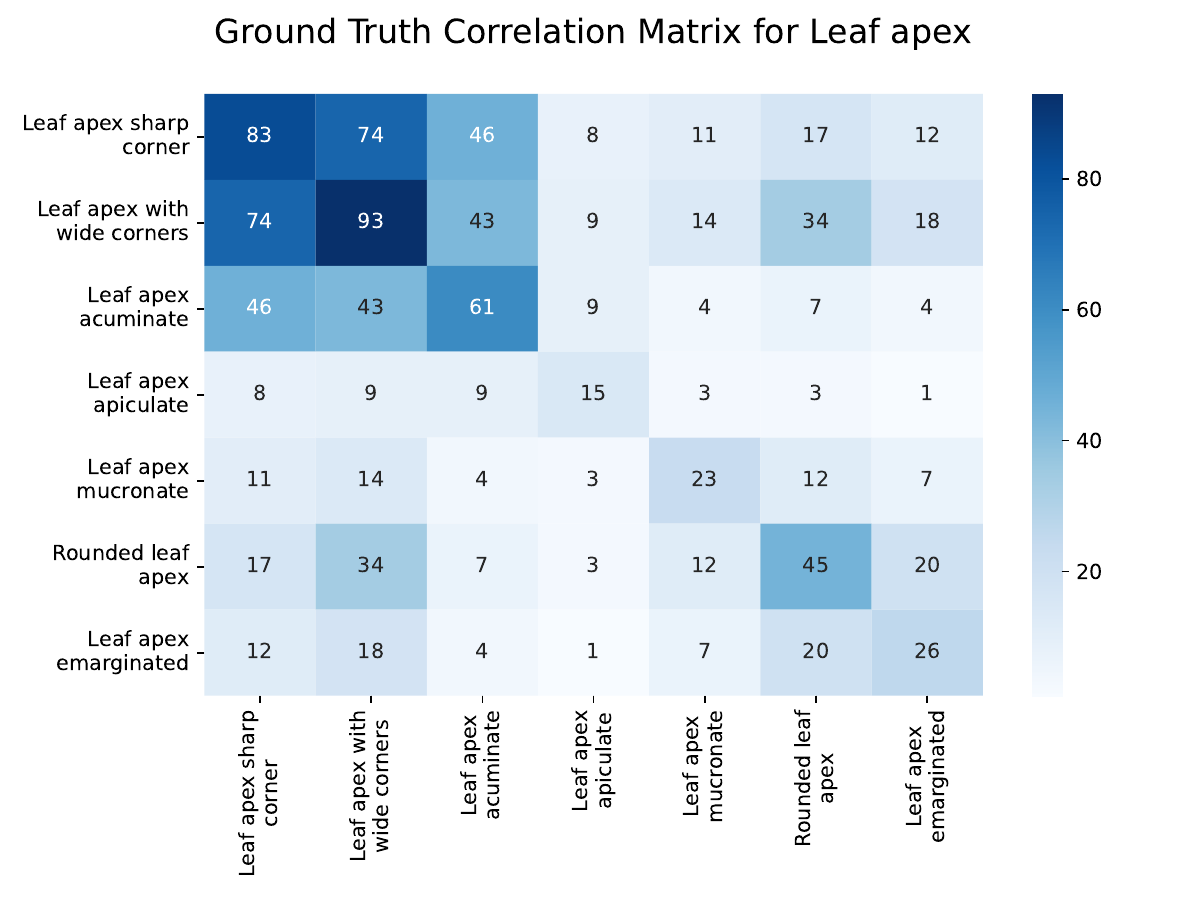}
         \caption{Leaf apex (W. African dataset) within the annotations.}
         \label{fig:Plantnet_leaf_apex_true}
     \end{subfigure}
     \begin{subfigure}[b]{0.49\textwidth}
         \centering
         \includegraphics[width=\textwidth]{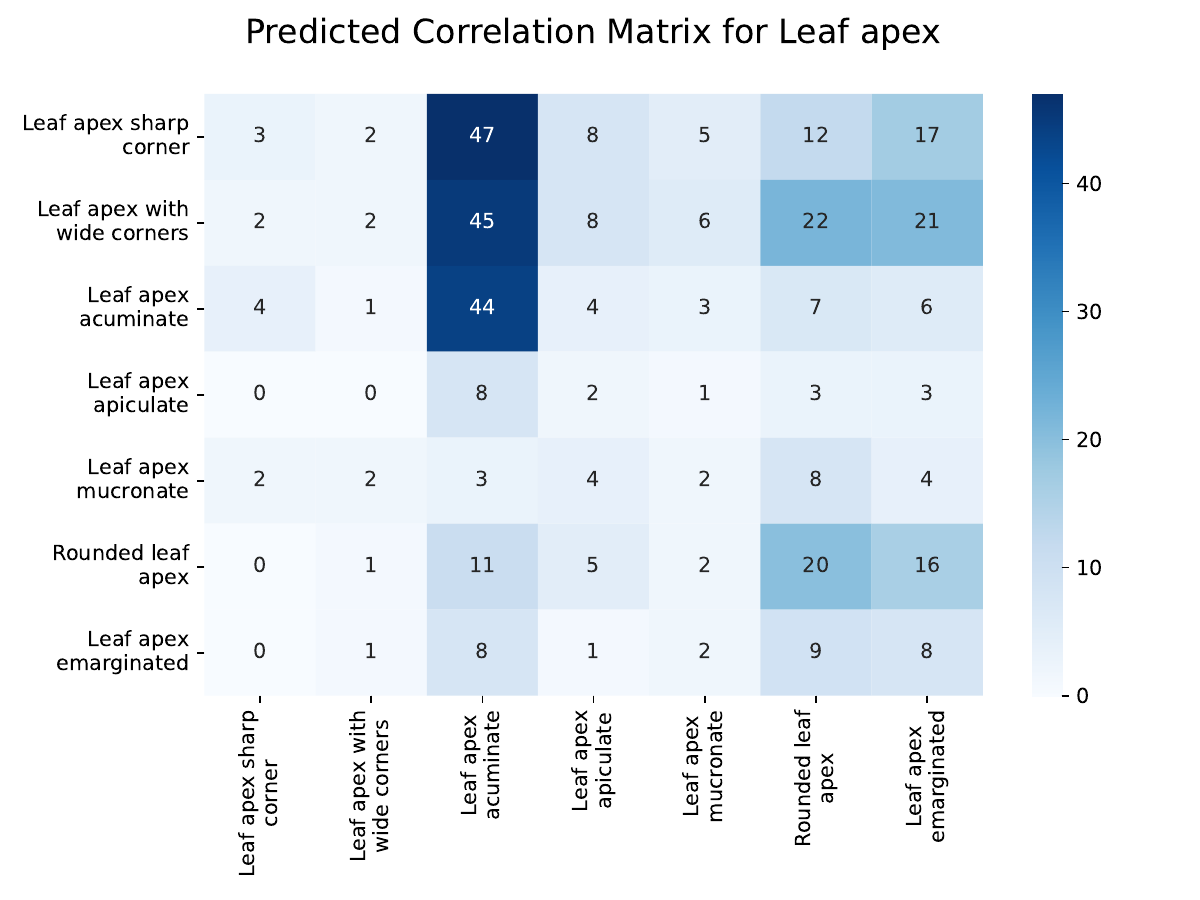}
         \caption{Leaf apex (W. African dataset) between predictions (columns) and annotations (rows).}
         \label{fig:Plantnet_leaf_apex_pred}
     \end{subfigure}
        \caption[Example co-occurrence matrices for worst and best performing traits.]{Co-occurrence matrices for \textit{stem shape} and \textit{Leaf apex} in the West Africa dataset, respectively the trait with the highest and the trait with the lowest F1 scores, i.e., 0.95 and 0.54 (left), along with the corresponding co-occurrences between the predictions and the annotated values (right). The co-occurrence patterns are conserved, except for \textit{Leaf apex sharp corner} and \textit{Leaf apex wide corners}, that are generally predicted as \textit{Leaf apex acuminate}. Note that these three trait values are highly correlated in the annotation.}
        \label{fig:multi_conf_3}
\end{figure}

In order to visualize the most typical mistakes the model commits on this multi-label task, in which more than one trait value is allowed per trait, we show two co-occurrence matrices side by side; one corresponding to the co-occurrences found within the annotated data (that is, which trait values are simultaneously present in a species) and a second one with the co-occurrences between the annotations and our predictions (which values are predicted for species that are annotated with a certain value).
Figure~\ref{fig:multi_conf_1} shows an example for the traits \textit{leaf position} and \textit{fruit type} in the Caribbean dataset and \textit{fruit} in the West African dataset. We can see that the general patterns of co-occurrence are maintained.
In addition, we can sometimes see that the committed confusions are often reasonable. For instance, if we look at the \textit{leaf position} trait, we see that our approach has returned \textit{opposite} when the manual annotations stated \textit{alternate-opposite}, \textit{opposite, whorls of 3} and \textit{opposite, whorls of 3, alternate}.
In Figure~\ref{fig:multi_conf_2} we display these confusion matrices for two traits in the Palms dataset: \textit{fruit size}, with two possible values, and \textit{fruit color}, with 12 possible values. In the latter we see that, although the correct values are often retrieved, the large number of options and potential ambiguities leads to a much larger number of false positives.
Finally, in Figure~\ref{fig:multi_conf_3} we show the two traits with the highest and lowest overall F$_1$ scores: \textit{stem shape} and \textit{leaf apex}, both from the West Africa dataset. We can see that the high scores in \textit{stem shape} are driven both by the fact that only two possible values are allowed and that it is a very imbalanced trait, with the vast majority of species having the same value.
On the other hand, \textit{leaf apex} not only has seven possible values, but they also show a very high overlap, which can be seen in the large off-diagonal values in three values of the annotations co-occurrence. Our pipeline tends to predict mostly one of these three: \textit{leaf apex accuminate}, while ignoring the other two.

\subsubsection*{Evaluation of the false negative rate}

In this section, we compare the ability of the LLM to predict ``NA'' in cases where no information about the desired trait can be found by comparing its responses to those of expert botanists on the same sentences.
This allows us to estimate whether the coverage rate actually corresponds to the actual data availability in the harvested text.
The confusion matrix in Table~\ref{tab:confusion_matrix_survey} shows that the LLM, in the setting used in this work, does not have a strong bias towards over- or under-detecting traits in text. Out of the 1216 text samples used in the survey, 24\% were deemed to contain relevant trait information by the botanists, while the LLM reported found traits in 22\%.
By comparing the responses between the LLM and the botanists, we obtain a macro averaged F$_1$ score of 0.82, with an F$_1$ score of 0.72 for the positive class and 0.92 for the negative class. 
The precision being higher than the recall suggests that the model has a conservative bias, with a tendency to under-report traits rather than hallucinating them.
Around 32\% of the traits that were reported as ``NA'' did actually contain information in the text that was missed by the LLM.
The observed precision is roughly in line with the performance of the approach, using the whole per-species text and set of traits in a single prompt, when compared to the manually curated species-trait matrix (Table~\ref{tab:traits_accuracy}, suggesting that the amount of input text provided in the prompt does not affect the quality of the results.

\begin{table}[h!]
\centering
\begin{tabular}{|c|c|c|c|}
\cline{2-4}
\multicolumn{1}{c|}{} & \textbf{LLM Missing} & \textbf{LLM Found} & \textbf{Total} \\ \hline
\textbf{GT Missing}   & 856                  & 62                 & 918            \\ \hline
\textbf{GT Found}     & 94                   & 204                & 298            \\ \hline
\textbf{Total}        & 950                  & 266                & 1216           \\ \hline
\end{tabular}
\caption[Confusion matrix of the trait presence prediction.]{Confusion matrix between the ground truth (GT) survey responses and the responses from the LLM. In this experiment, for the LLM queries, instead of using only the known ground truth values for each trait, we consider all its possible values. If there is evidence of any value present in the sentence, we consider the value found, otherwise missing. At the same time, for the responses found in the surveys, we consider the value found only when the reviewers responded ``Can infer correct Value'' and missing otherwise.}
\label{tab:confusion_matrix_survey}
\end{table}

\subsubsection*{Additional experimental results}

In order to further investigate the impact of some of the design choices, we evaluate on the Caribbean dataset with two additional LLM settings.
First, we want to verify that querying the LLM with all traits simultaneously in a single prompt does not substantially degrade the results.
In Table~\ref{tab:additional_results}, we observe that querying a single trait at a time results in a mere half percent point improvement, while the number of input token required is over an order of magnitude larger, from 150k input tokens to 1.67M, due to having to provide the input text and instructions as many times per species as there are traits.
The main improvement is on coverage, that goes from 55\% to almost 58\%.
Note that running our experiments with all traits at once on the over 700 plant species that comprise the three datasets has only required around \$30 in Mistral AI credits.
Second, we evaluate the applicability of the open source model Mixtral-8x22B, which was released after we had run the main set of experiments.
The results in Table~\ref{tab:additional_results} show that this model is able to provide comparable results to mistral-medium, with only a 0.4\% lower F$_1$, and an improved coverage of over 60\%.

\begin{table}[h!]
    \centering
    \begin{tabular}{l|c|c|c||c}
        \hline
         
          Model / setting  & Precision & Recall & F$_1$ & Coverage\\\hline
          mistral-medium / all traits& 0.7493 & 0.7800 & 0.7643 & 0.5500\\ 
          mistral-medium / single trait & 0.7507 & 0.7920 & 0.7708 & 0.5791\\
          Mixtral-8x22B / all traits & 0.7519 & 0.7726 & 0.7602 & 0.6052\\
          GPT-3.5 Turbo / all traits & 0.6665 & 0.7390 & 0.7009 & 0.3031
          \\\hline
          
    \end{tabular}
    \caption[Trait prediction metrics with alternative settings.]{Results with four different settings. Top: using the mistral-medium model with a single prompt per species, querying for all traits simultaneously. Second: mistral-medium model querying for a single trait in each prompt. Third: Mixtral-8x22B model with all traits in a single prompt. Bottom: GPT-3.5 Turbo model with all traits in a single prompt. Precision, recall, and F$_1$ score are calculated with respect to the manually annotated Caribbean dataset, along with the coverage, \emph{i.e.}, the proportion of trait-species entries for which at least one value is found. The accuracy metrics are computed only for these entries.}
    \label{tab:additional_results}
\end{table}

\section*{DISCUSSION}

Our study reveals that while the majority of species yield useful descriptive text, a significant portion of the Palms dataset provided no sentences at all. This issue is partly attributable to the study's focus on English-language HTML websites. 
This could be mitigated with the inclusion of non-English text, as many botanical descriptions are available in local languages. Expanding the language scope would be particularly feasible for languages with a significant online presence and a history of botanical use, such as French, Spanish, and Portuguese.
Broadening the range of structured online resources used to train the description detector is one alternative for this multilingual expansion. Although modern LLMs are multilingual, the most significant challenge lies in extending our descriptive text detection approach to multiple languages. This could be addressed either by training on structured websites in various languages or by employing a higher capacity LLM in this step via in-context learning~\cite{NEURIPS2020_1457c0d6}, which would leverage the multilingual nature of most LLMs. 
With this last approach, a few examples of descriptions or a definition of the characteristics of descriptions could suffice, removing the need to train on a large set of multilingual description examples.
The markedly lower recall compared to precision in detecting descriptive text is expected, given the nature of the data and the loss function used, which accounts for a considerable amount of label noise. The low recall suggests that the model often determined that nearly half of the text within descriptive sections did not genuinely pertain to descriptions. While this may lead to the omission of some relevant sentences, it also results in a more concise and focused corpus.

Our quantitative trait extraction results show that the proposed pipeline is able to return a value for over half of the traits in the three considered species-trait matrices, with an average F$_1$-score of over 0.75.
In addition, the inspection of the errors it commits suggests that they tend to be relatively reasonable mistakes, with similar trait values being typically confused for one another.
The results on the false negative rate evaluation show that, in general, the LLM is well-balanced and has no strong tendency towards either hallucinating nor ignoring information.
The fact that using a single trait per query results in very similar performance is a sign that this behavior does not depend on the number of simultaneously queried traits.
These results mean that the low average coverage rate, of about 55\%, can probably be blamed on a lack of information in the harvested dataset, rather than on the LLM being unable to pick up the information. A focus on improving the amount of textual information would, therefore, be the best way of further improving the trait coverage.
Nonetheless, we have directly used the trait and trait value names as they were proposed by the original authors of the species-trait datasets. 
It is likely that these specific formulations are not the best possible for our task and can thus be optimized for prompting, such as by including descriptions of the traits and their possible values.
In addition, the results using Mixtral-8x22B show that, at the time of publication, it would be possible to reproduce the results in this paper, and scale the approach to new species, using a model with openly available weights.  

This study focused on a relatively small number of plant species, approximately 700 in total, for which manually curated trait data was available for evaluation. We attempted to mitigate geographic bias toward Europe and North America by exclusively considering tropical species, which are more likely to be representative of the data availability for the World flora than these overrepresented regions~\cite{kattge2020try}. However, the study was limited to woody plants. This constraint may affect the generalizability of our findings to the global flora.
We also observed that our approach successfully filled in only a little over half of the traits, with up to 25\% of species in the Palms dataset failing to yield any text during the Web crawling phase, thus resulting in no identified traits.
The primary limitation of our method lies in its reliance on species and traits that are textually documented online. As a result, the approach is more suited to retrieving morphological traits, which are frequently used in plant species descriptions that can be found on the Web.
This contrasts with existing trait database initiatives, such as TRY~\cite{kattge2011try}, BIEN~\cite{maitner2018bien} or TraitBank~\cite{caldwell2014using}, which contain traits based on measured specimens, without a focus on the morphological traits that are typically used for species description and identification.
In addition, changes in taxonomic nomenclature can lead to missing valid information that uses an outdated name.
To address this limitation, the procedure could be enhanced by incorporating less stringent filtering during text harvesting and including the use of synonyms. The capacity of the pipeline to capture more text could be further improved by implementing compatibility with JavaScript-based websites and PDF documents~\cite{folk2023floratraiter}.
Finally, our study focused on categorical traits, though we believe the approach could be adapted for other types of trait formulations, such as numerical values, with modifications to the LLM prompt, along with an additional step to deal with different measuring units. We plan to explore this possibility in future work.

\subsection*{Concluding remarks}

We develop and evaluate a pipeline that leverages recent advances in large language models to extract trait information for any set of species from unstructured online text. 
Unlike other recent approaches that require species-trait information for training~\cite{domazetoski2023using} or manual curation~\cite{folk2023floratraiter}, ours does not require any manual annotations for training nor any curation step. The only manual effort required is the initial creation of the list of traits and the possible trait values along with the list of species name to be examined; this means  that the trait extraction can be effortlessly scaled to new sets of species without the need for previous knowledge on species-trait relations.
These results point towards the potential of this type of methodology for leveraging the large amounts of unstructured text data available online on species descriptions.
Although in this work we limited the list of traits to those present in the reference, hand-crafted datasets, we could adapt the approach to use for more general lists~\cite{castellan2023back} in order to allow scaling up to much larger floras.

\section*{Data availability statement}

All the code and data needed to reproduce the results in this paper are available at \url{https://github.com/konpanousis/AutomaticTraitExtraction}.
The version used for the results in this paper, along with all the associated data, can be found in \url{https://doi.org/10.5281/zenodo.13969765}.
In addition to providing the code, we also provide an easy to run version, with a working example, in the form of a Python Jupyter notebook. It is directly runnable on Google Collab without the need for specialized hardware.

\section*{Author contributions statement}
D.M., R.V., I.A., P.B., A.J, H.G. and D.K. conceived the experimental setup,  K.P., C.L, R.V and D.M. conducted the experiments, A.P., P.B. and D.K. provided the data, K.P., C.L. and D.M. analysed the results.  All authors reviewed the manuscript. 

\section*{Acknowledgements}
We would like to thank Vanessa Hequet for providing expert annotations on trait information presence that were used in this work.\\
The research described in this paper was partially funded by the European Commission via the GUARDEN project, which have received funding from the European Union’s Horizon Europe research and innovation programme under grant agreements 101060693.

\section*{In memoriam}

This work is dedicated to the memory of Robert van de Vlasakker (May 23$^\text{rd}$ 1992 --- May 27$^\text{th}$ 2023), who was the first to initiate this research and whose dedication and passion laid the foundation for this project. As a student, he showed exceptional promise and maturity, and made significant contributions to the early stages of this work. Although Robert sadly passed away before the completion of the manuscript, his enthusiasm, vision and extraordinary resilience remain an inspiration to all of us. We want to honor his memory with the publication of this piece of research, which would not have been possible without his pioneering efforts.

\newpage

\bibliographystyle{apalike}
\bibliography{sample}

\end{document}